%% file: colm2026_conference.tex
\documentclass{article} 
\usepackage[preprint]{colm2026_conference}

\usepackage{microtype}
\usepackage[backref=page]{hyperref}
\usepackage{url}
\usepackage{booktabs}
\usepackage{graphicx}
\usepackage{lineno}
\usepackage{amsmath}
\usepackage{xcolor}
\usepackage{colortbl}
\usepackage{multirow}
\usepackage{wrapfig}
\usepackage{pgf}

\definecolor{darkblue}{rgb}{0, 0, 0.5}
\hypersetup{colorlinks=true, citecolor=darkblue, linkcolor=darkblue, urlcolor=darkblue}

\title{Where does output diversity collapse in post-training?} 

\author{Constantinos Karouzos \qquad Xingwei Tan \qquad Nikolaos Aletras \\[1em]
School of Computer Science \\
University of Sheffield, UK \\
\texttt{\{kkarouzos1, xingwei.tan, n.aletras\}@sheffield.ac.uk}}

\begin{document}

\ifcolmsubmission
\linenumbers
\fi

\maketitle

\begin{abstract}
Post-trained language models produce less varied outputs than their base counterparts. This output diversity collapse undermines inference-time scaling methods that rely on varied samples, and risks homogenizing model outputs on creative and value-laden tasks. Prior work attributes collapse to specific post-training methods, without separating the role of training data composition from the method, or the generation format from the model weights. We trace output diversity through three parallel post-training lineages of Olmo 3, Think (chain-of-thought distillation), Instruct (broad multi-source data), and RL-Zero, across 15 tasks and four text diversity metrics. We find that the location of collapse co-varies with data composition: the Think lineage loses most semantic diversity at supervised fine-tuning, and the effect of DPO is larger in Instruct than in Think. Suppressing chain-of-thought reasoning at inference in Think models drops accuracy on hard tasks, yet leaves answer-level diversity unchanged, showing that the collapse is embedded in the model weights by training data, not imposed by the generation format. Decomposing diversity loss on six verifiable tasks into a quality-control component (removal of incorrect outputs) and a residual component (genuine narrowing among correct outputs) reveals that the split is task-dependent, and Think models retain more correct-answer diversity than Instruct despite collapsing more in aggregate. Our results indicate that diversity collapse is determined during training by data composition and cannot be addressed at inference time alone.\footnote{Code: \url{https://github.com/ckarouzos/where-diversity-collapses/}}
\end{abstract}

\section{Introduction}
\label{sec:intro}

Large language models (LLMs) rely on post-training to improve helpfulness, safety, and instruction compliance. Post-training combines supervised fine-tuning~\citep[SFT;][]{ouyang2022traininglanguagemodelsfollow} on curated demonstrations, and direct preference optimization~\citep[DPO;][]{rafailov2023direct} or reinforcement learning from human feedback (RLHF). However, this results in output diversity collapse, i.e., models produce more uniform outputs than their base counterparts across summarization~\citep{kirk2024understanding}, reasoning~\citep{dang2025diversity}, and open-ended generation~\citep{jiang2025hivemind}.
Diversity collapse limits self-consistency~\citep{wang2023selfconsistency}, pass@$k$ sampling~\citep{chen2021evaluatinglargelanguagemodels}, and test-time compute scaling~\citep{snell2025scaling}. \citet{kamigaito2025diversity} show diversity is the mechanism underlying inference scaling laws. The algorithmic causes are well-understood~\citep{wang2024beyond, ma2025gradient, gxchen2025kl}, yet diversity collapses across task types. This leads LLMs to produce less diverse outputs than a basic web search~\citep{wright2025epistemic}, co-writing with LLMs reduces content diversity~\citep{padmakumar2024writing}, and single-reward RLHF can amplify majority preferences to near-total dominance~\citep{chakraborty2024maxmin}.

Yet, prior work attributes collapse to specific algorithms. DPO in narrative generation~\citep{peeperkorn2025mindgapconformativedecoding}, the reward step in creative tasks~\citep{omahony2024attributing}, and SFT in reasoning~\citep{dang2025diversity}, without investigating the effect of \emph{data} compositions. \citet{ma2025reasoning} suppress chain-of-thought~\citep[CoT;][]{wei2022chain} at inference but measure only accuracy, not diversity. No existing study isolates the role of the training \emph{method} from the training \emph{data}, or the generation \emph{format} from the model weights.

Two questions remain open: (1) \textit{does the diversity collapse co-vary with the post-training method or with the post-training data composition,} and (2) \textit{does the CoT format itself constrain diversity at inference, or is the collapse embedded in the model weights?} 

\begin{figure*}[t]\centering
\vspace{-25pt}
\includegraphics[width=0.85\columnwidth]{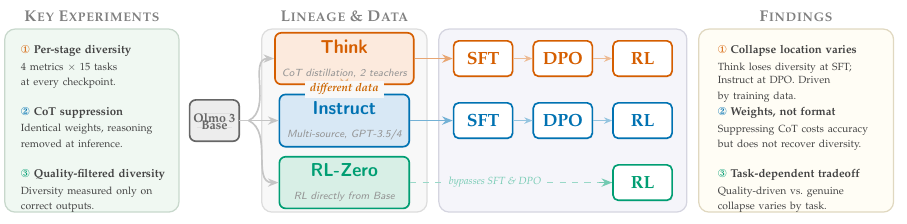}
\caption{Study design. We trace output diversity through three parallel post-training lineages of Olmo 3, to identify where, why, and how much diversity is lost.}
\vspace{-10pt}
\label{fig:intro}
\end{figure*}

We answer these questions through a controlled experimental setting (Figure~\ref{fig:intro}). We monitor the output diversity of the open weight and data Olmo 3 model family~\citep{olmo2025olmo3}, which releases checkpoints of all post-training stages across three parallel lines. \textbf{Think} and \textbf{Instruct} variants share the same post-training recipe (SFT$\to$DPO$\to$RL) but differ in data, while \textbf{RL-Zero} bypasses SFT and DPO entirely. Evaluating 13 models across 15 tasks with four diversity metrics, we show that the same post-training method produces different diversity outcomes depending on the upstream data composition, and that each stage plays a distinct role. Our contributions:
\begin{itemize}
    \item We compare Think vs.\ Instruct lineages, showing that collapse location depends on data: narrow CoT distillation for Think models is associated with a larger drop at SFT, while the DPO drop is larger in Instruct models (\S\ref{sec:cliffs});
    \item We evaluate Think models with CoT suppressed at inference and find no diversity recovery on any task--stage combinations, while quality drops. Diversity collapse resides in the model weights, not in the CoT generation format (\S\ref{sec:nocot});
    \item We decompose diversity reduction into a quality-control component (removal of incorrect outputs) and a residual component (genuine narrowing among correct outputs), showing the split is task-dependent (\S\ref{sec:tradeoff}).
\end{itemize}

\section{Related work}
\label{sec:related}

\textbf{The reliability--diversity tradeoff in post-training.}
\citet{jiang2025hivemind} show that aligned models exhibit high output homogeneity across a wide range of model families and scales. \citet{kirk2024understanding} find that RLHF reduces both per input and across input diversity. Human co-writing with aligned models reduces content diversity~\citep{padmakumar2024writing}, and users brainstorming with ChatGPT produce less semantically distinct ideas~\citep{anderson2024homogenization}. In reasoning, SFT improves pass@1 but degrades pass@$k$~\citep{dang2025diversity}; base models outperform RLVR-trained models at large sample budgets~\citep{yue2025does}, and base models produce more diverse outputs~\citep{west2025base}. \citet{peeperkorn2025mindgapconformativedecoding} identified DPO as the steepest drop. \citet{karouzos2026empirical} show that under domain shift the adaptation strategy dominates the alignment objective. Current methods cannot selectively preserve diversity where it is beneficial~\citep{jain2025task}. Quality-adjusted diversity shows that preference-tuned models retain higher diversity among high-quality outputs~\citep{shypula2025evaluating}, and multi-dimensional linguistic benchmarks find that larger models are often less diverse than smaller ones~\citep{guo-etal-2025-benchmarking-linguistic}. Automatic diversity metrics lag behind human judgments~\citep{tevet-berant-2021-evaluating}, and sampling temperature cannot recover training-induced loss~\citep{verine2025improving}. 

\textbf{Mechanisms and mitigations.}
DPO's gradient imbalance suppresses dispreferred responses~\citep{ma2025gradient}, and likelihood displacement shifts probability to unintended outputs~\citep{razin2025unintentional}. 
KL-regularized RL specifies unimodal targets by construction~\citep{gxchen2025kl}, preference collapse arises from KL amplification~\citep{xiao2024preference}, and chat templates induce diversity collapse~\citep{yun-etal-2025-price}. Training on recursively generated synthetic data causes progressive tail disappearance~\citep{shumailov2024model}. Proposed mitigations include forward-KL optimization~\citep{wang2024beyond}, entropy-constrained RL~\citep{pan2026qempo}, decoupled regularization~\citep{slocum2025diverse}, game-theoretic SFT~\citep{li2025preserving}, diversity-aware preference optimization~\citep{li2025darling, lanchantin2025divpo}, and conformative decoding~\citep{peeperkorn2025mindgapconformativedecoding}. A single reward function is insufficient to represent diverse human preferences~\citep{chakraborty2024maxmin}.

\section{Experimental setup}
\label{sec:setup}

\subsection{Models and training lineages}
\label{sec:background}

We study 13 Olmo~3 checkpoints at the 7B scale. Post-training applies up to three stages, SFT, DPO, and RL, starting from the same base model. 

\textbf{Base} (1 model). The base model is pretrained on Dolma~3 Mix (6T tokens), midtrained on Dolmino Mix (100B tokens), and context-extended to 65K tokens.

\textbf{Think} (3 models: Think-SFT, Think-DPO, Think). SFT trains on ${\sim}$2.3M synthetic CoT~\citep{wei2022chain} reasoning traces using (prompt, completion) pairs from two teachers: QwQ-32B~\citep{qwq32b} and DeepSeek-R1~\citep{deepseek2025r1}.  DPO uses ${\sim}$200K Delta Learning~\citep{geng2025deltalearning} pairs. The RL stage uses a variation of GRPO~\citep{shao2024deepseekmath} with verifiable rewards and no KL penalty, and trains on ${\sim}$105K prompts, to produce Think. 

\textbf{Think-not-thinking.} To isolate the contribution of the CoT generation format from the learned weights, we additionally evaluate all three Think checkpoints with CoT suppressed by prefilling an empty \texttt{<think>$\backslash$n</think>$\backslash$n} block, forcing direct answers.

\textbf{Instruct} (3 models: Instruct-SFT, Instruct-DPO, Instruct). SFT \emph{initializes from} Think-SFT, then trains on ${\sim}$2.2M examples that include function-calling, strip reasoning traces, and draw from multiple sources (GPT-3.5, GPT-4, GPT-4.1;~\citealp{openai2023gpt4}) rather than two teachers. DPO (${\sim}$260K pairs) uses the same pool of prompts as Think-DPO but with the thinking mode disabled, adding multi-turn and GPT-judged preference pairs. The same RL stage as Think produces the final Instruct model.

\textbf{RL-Zero} (6 models). Applies RL training directly to Base, bypassing SFT and DPO. Four Olmo~3 variants target different reward domains: RL-Zero-Math, RL-Zero-Code, RL-Zero-IF, and RL-Zero-General (${\sim}$105K prompts each). Two additional Olmo~3.1 variants (RL-Zero-Math$^{3.1}$, RL-Zero-Code$^{3.1}$) are trained for more steps.

\subsection{Tasks and Data}
\label{sec:tasks}

\textbf{Summarization.} TL;DR~\citep{volske-etal-2017-tl}, CNN/DailyMail~\citep{nallapati-etal-2016-abstractive}, and XSum~\citep{narayan-etal-2018-dont}. Bounded output length controls for length confounds, and multiple valid summaries provide a clear diversity signal.

\textbf{Code.} HumanEval~\citep{chen2021evaluatinglargelanguagemodels}, MBPP~\citep{austin2021programsynthesislargelanguage}, and CRUXEval~\citep{gu2024cruxeval}. Outputs can be syntactically different but functionally identical, and RL directly optimizes code tasks.

\textbf{Reasoning.} GSM8K~\citep{cobbe2021trainingverifierssolvemath}, MATH-Algebra, MATH-Geometry~\citep{hendrycks2021measuring}, and TruthfulQA~\citep{lin-etal-2022-truthfulqa}, the primary Think and RL-Zero training domain. Diversity here measures variation in solution \emph{strategy} with answers held constant.

\textbf{Instruction following.} Alpaca~\citep{alpaca}, open-ended, and IFEval~\citep{zhou2023ifeval}, with verifiable format constraints.

\textbf{Creative writing.} WritingPrompts~\citep{fan-etal-2018-hierarchical}, where diversity is intrinsically desirable.

\textbf{Value pluralism.} PRISM~\citep{kirk2024prism} and WildBench~\citep{lin2025wildbench}, which test whether alignment imposes a single perspective on contested topics.

We measure training--evaluation overlap using $C_{13}$ 13-gram matching~\citep{lambert2025tulu} between the four Dolci post-training datasets and all fifteen evaluation tasks (Appendix~\ref{app:decontamination}). Nine datasets show negligible overlap (${\le}\,2\%$). HumanEval, CRUXEval, IFEval, MATH-Algebra, MATH-Geometry, and WildBench show elevated overlap (7--30\%), traceable to shared upstream data. While we flag these benchmarks, our findings on contaminated tasks are consistent with the patterns on the clean tasks.

\subsection{Metrics}
\label{sec:metrics}

We measure diversity along four complementary axes (detailed definitions in Appendix~\ref{app:metric_details}). \textbf{EAD}~\citep{liu-etal-2022-rethinking} counts unique $n$-grams normalized against the expected count under a uniform draw (averaged over $n \in \{1,\dots,5\}$), capturing \emph{lexical} diversity. \textbf{SBERT} computes mean pairwise cosine distance of sentence embeddings~\citep[\texttt{all-mpnet-base-v2};][]{reimers-gurevych-2019-sentence}, capturing \emph{semantic} diversity (0~= collapse, 1~= dissimilar). For code tasks we additionally report \emph{semantic} diversity with UniXcoder~\citep{guo2022unixcoder} embeddings (Appendix~\ref{app:code_diversity}). \textbf{NLI} scores output pairs with an NLI classifier~\citep[\texttt{roberta-large-mnli};][]{liu2019robertarobustlyoptimizedbert}, following~\citet{stasaski-hearst-2022-semantic}, capturing \emph{logical} diversity; code tasks are excluded. \textbf{Vendi Score}~\citep{friedman2023the} measures the effective number of dissimilar outputs via eigenvalue entropy of the SBERT similarity kernel (VS${=}1$: identical, VS${=}K$: orthogonal). For code-generation tasks we also report \textbf{AST subtree diversity}, the mean pairwise Jaccard distance on AST subtree multisets ~\citep{shypula2025evaluating}, on correct outputs only (Appendix~\ref{app:code_diversity}).

\textbf{Quality.} For the six tasks with verifiable answers (GSM8K, MATH-Algebra, MATH-Geometry, HumanEval, MBPP, IFEval), we report: accuracy@1 (greedy decoding), majority vote@16 (most frequent answer among $K{=}16$ samples), and pass@16 (at least one correct among $K$). For code tasks we use the unbiased pass@$k$ estimator. For IFEval we report strict and loose constraint satisfaction. For the eight tasks without verifiable answers we evaluate quality using LLM-as-judge (\texttt{gpt-4.1-mini}) with established protocols (Appendix~\ref{app:quality}).

\textbf{Quality-filtered diversity.} We decompose diversity into a quality-control component (removal of incorrect outputs) and a residual component (genuine narrowing among correct outputs). $D_a$ (SBERT on all $K$ outputs) and $D_c$ (SBERT on the $K_c \geq 2$ correct outputs). The gap $D_a - D_c$ reflects diversity from error variety; $D_c$ captures genuine narrowing among correct solutions. We report analogous Vendi scores $V_a$ and $V_c$. 

For each model--task pair, we generate $K{=}16$ outputs per prompt at $T{=}0.6$, top-$p{=}0.95$. Base recommends $T{=}1.0$; we use matched settings for controlled comparison (Appendix~\ref{app:temperature}). For all Think-lineage models, we strip \texttt{<think>...</think>} reasoning traces before computing any metric, so that all diversity and quality scores reflect the \emph{final answer} only. Implementation details are in Appendix~\ref{app:implementation}.

\section{Results}
\label{sec:results}

\begin{figure*}[t]
\centering
\vspace{-25pt}
\includegraphics[width=0.85\textwidth]{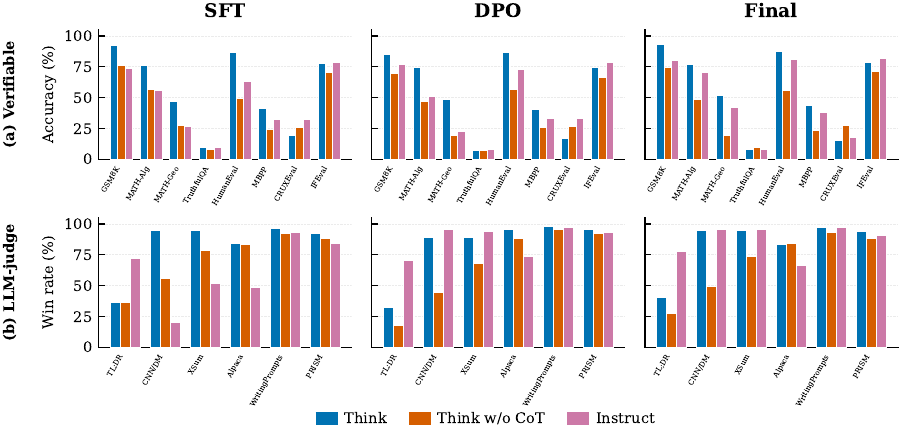}
\caption{Quality of generations for Think, Think-not-thinking, and Instruct, across stages. \textbf{Top}: accuracy on eight verifiable tasks. \textbf{Bottom}: LLM-judge win rates on six tasks.}
\label{fig:quality}
\vspace{-15pt}
\end{figure*}

We present results around three questions. First, \emph{where} does diversity collapse along each lineage (\S\ref{sec:cliffs}; Figure~\ref{fig:lineage_stages}, Table~\ref{tab:stages})? Second, does the CoT generation format itself constrain diversity (\S\ref{sec:nocot}; Figures~\ref{fig:quality}--\ref{fig:wildbench})? Third, how much of the observed collapse is attributable to quality control (\S\ref{sec:tradeoff}; Figures~\ref{fig:qfd_decomposition}--\ref{fig:mv_gain})?

\subsection{Lineage-dependent diversity collapse}
\label{sec:cliffs}

\begin{figure*}[t]
\centering
\vspace{-20pt}
\includegraphics[width=\textwidth]{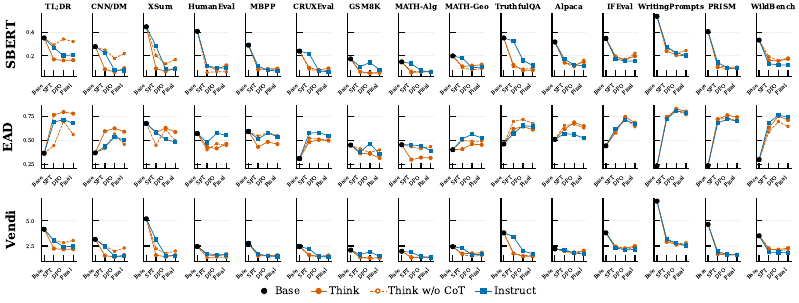}
\caption{SBERT, EAD, and Vendi Score across post-training stages. Think (orange) collapses at SFT; Instruct (blue) at DPO. Think w/o CoT (hollow) tracks Think.}
\label{fig:lineage_stages}
\vspace{-15pt}
\end{figure*}

\textbf{SFT asymmetry.} Think and Instruct share the same three-stage post-training, yet collapse at different stages. Think-SFT loses 62\% (Table~\ref{tab:stages}) of Base diversity on average, 24\% more than Instruct-SFT (38\%), uniformly across all 15 tasks, consistent with \emph{completion homogeneity} from two teachers rather than prompt overlap. This challenges findings of minimal SFT impact on diversity~\citep{guo-etal-2025-benchmarking-linguistic} and suggests that the effect depends on the breadth of the SFT data. Collapse magnitude also scales with task difficulty (Figure~\ref{fig:lineage_stages}). Think-SFT retains only 36\% of Base diversity on GSM8K (92\% accuracy) but 54\% on MATH-Geometry (50\% accuracy). Easier tasks with a dominant solution strategy collapse the most. Instruct-SFT, despite initializing from the already-collapsed Think-SFT, recovers a median 40\% of the lost diversity, likely due to its multi-source data. As Instruct-SFT initializes from Think-SFT, this recovery also reflects the dynamics of retraining a collapsed model.

\input{tables/stage_attribution}

\textbf{DPO asymmetry.} DPO erases more diversity in Instruct than in Think, as Think has already collapsed at SFT, leaving little for DPO to remove. The effect is largest on summarization and code-reasoning tasks, where Instruct-SFT had preserved substantial diversity. On three math/code tasks, Think-DPO actually \emph{increases} diversity slightly, and Instruct-DPO does the same on GSM8K, suggesting that DPO can partially correct a collapsed SFT distribution.

\begin{wrapfigure}{r}{0.4\columnwidth}
\vspace{-15pt}
\centering
\includegraphics[width=0.4\columnwidth]{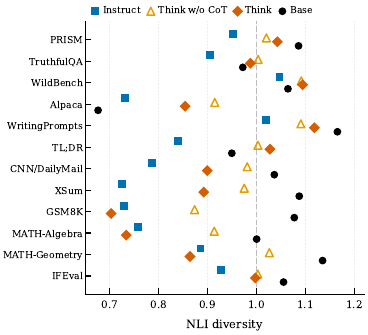}
\vspace{-15pt}
\caption{NLI diversity.}
\label{fig:nli_sbert}

\end{wrapfigure}

\textbf{RL reversal.} Think's RL stage increases semantic diversity on most tasks, primarily code and summarization. The recovery is modest (roughly 5\% of total diversity lost) but directionally consistent. Both lineages use the same RLVR method, so the asymmetry likely reflects the input state: Think enters RL already at its diversity floor, leaving room for exploration, while Instruct enters with residual diversity that RL continues to compress. On GSM8K, Instruct RL erases 37\% of Base diversity, the largest single-stage loss outside SFT, as the verifiable reward concentrates probability on the dominant correct strategy. The RLVR stage also produces lexically \emph{more uniform} outputs (EAD decreases on nearly all tasks), suggesting it standardizes surface form while broadening semantic content.

\textbf{Convergence.} RL-Zero bypasses both bottlenecks (Figure~\ref{fig:lineage_stages}), retaining $\ge71\%$ of Base diversity (median 94\%). Both supervised lineages converge to similar final diversity floors (with Think slightly higher on 11/15 tasks), despite different trajectories: data composition co-varies with \emph{when} and \emph{how sharply} diversity is lost. Table~\ref{tab:stages} summarizes the stage-wise attribution. Full per-task breakdowns are in Appendix~\ref{app:stage_attribution}.

The collapse is semantic, not lexical (Figure~\ref{fig:lineage_stages}). Per input SBERT drops from 0.32 (Base) to 0.12 (Think) and 0.11 (Instruct), and the Vendi Score drops from ${\sim}$3.4 effective modes to ${\sim}$1.8 (final), with near-total collapse on math (GSM8K: 1.3 modes, MATH-Algebra: 1.4), 16 samples carry essentially no more semantic diversity than one. EAD (Figure~\ref{fig:lineage_stages}) remains stable or \emph{increases}, even as semantic diversity drops. Aligned models use varied vocabulary and phrasing to express semantically identical content. Think's EAD on WritingPrompts rises from 0.23 to 0.80, while SBERT falls from 0.54 to 0.20, a pattern replicated across open-ended tasks. For natural language tasks, NLI diversity (Figure~\ref{fig:nli_sbert}) drops on most tasks, though the gap varies. Post-trained models still make logically distinct claims. The gap is largest for Think models, where CoT reasoning preserves logical structure even as the surface distribution narrows.

Value-pluralism tasks suffer the steepest Think collapse (PRISM $-78\%$, TruthfulQA $-79\%$), as narrow two-teacher distillation cannot represent the range of perspectives these tasks require. On PRISM, Think's NLI (Figure~\ref{fig:nli_sbert}) scores remain above 1.0 (net contradictions), meaning the model still samples contradictions despite converged phrasing, though we cannot determine whether this is genuine stance plurality or internal incoherence. Instruct drops NLI below 1.0, indicating homogenization of both form and stance (Figure~\ref{fig:nli_sbert}). Think's NLI remains above the contradiction threshold on value-pluralism and creative tasks where Instruct's drops below. 
Creative writing (WritingPrompts) shows the highest Base diversity (6.9 Vendi modes) and the sharpest quality--diversity tension. Think and Instruct both collapse to ${\sim}$0.20 SBERT and ${\sim}$2.6 modes ($-63\%$), yet achieve ${>}$97\% pairwise win rate against Base, producing better stories at the cost of formulaic variation. RL-Zero retains ${\sim}$100\% of Base diversity, but wins only ${\sim}$50\%, consistent with the absence of a creative-writing reward signal. NLI diversity remains above 1.0 for all models on WritingPrompts (Think 1.12, Instruct 1.02, RL-Zero 1.15), meaning post-trained models still produce logically distinct narratives despite semantic convergence. Full per-task breakdowns are in Appendix~\ref{app:tables}.

\subsection{Think-not-thinking: CoT as reliability, not diversity}
\label{sec:nocot}

\begin{wrapfigure}{r}{0.4\columnwidth}
\vspace{-15pt}
\centering
\includegraphics[width=0.4\columnwidth]{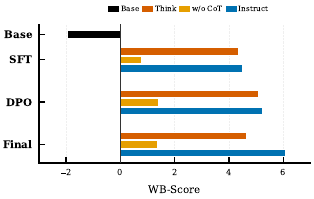}
\vspace{-15pt}
\caption{WildBench Score.}
\label{fig:wildbench}
\vspace{-1pt}
\end{wrapfigure}

Think and Instruct differ in both training data \emph{and} generation format. Think generates CoT reasoning traces before answering, while Instruct answers directly. To isolate the format's contribution, we evaluate all three Think models with CoT suppressed, we refer to these models as \emph{Think-not-thinking}. This is an out-of-distribution intervention, so we interpret the results as testing whether format removal recover diversity. Across tasks (Figure~\ref{fig:lineage_stages}), removing CoT \textbf{ does not recover diversity}. Think-not-thinking SBERT diversity matches Think, and Instruct shows similarly collapsed diversity. This holds at every stage (SFT, DPO, RLVR) and across every task category. IFEval shows a small increase ($+0.025$ SBERT), but this is modest relative to the Base-to-Think gap ($-0.153$).

CoT suppression \emph{does} affect accuracy (Figure~\ref{fig:quality}), with harder tasks losing more: IFEval $-8\%$, GSM8K $-18\%$, MBPP $-20\%$, MATH-Algebra $-28\%$, HumanEval $-32\%$, MATH-Geometry $-32\%$. The quality cost is task-dependent (Figure~\ref{fig:quality}), CoT suppression is negligible for open-ended generation (no change for Alpaca, WritingPrompts $-4\%$) but severe for summarization (CNN/DM $-48\%$) and complex helpfulness (WildBench Score $4.6\to1.4$, Figure~\ref{fig:wildbench}). In no case does suppression recover diversity. CoT improves reliability by helping the model execute its learned strategy, especially on hard problems, without broadening the answer-level diversity distribution. The output distribution is equally collapsed whether the model reasons explicitly or answers directly. One exception is WritingPrompts, where removing CoTs slightly \emph{increases} SBERT diversity ($+0.046$), suggesting that CoT imposes implicit narrative templates that constrain story generation. NLI diversity reveals a subtler pattern on math tasks: Think-not-thinking produces \emph{higher} NLI scores than Think (GSM8K: 0.87 vs.\ 0.70; MATH-Algebra: 0.91 vs.\ 0.73), despite identical SBERT. Without CoT, final answers are semantically collapsed but logically less entailing. The model generates diverse wrong answers rather than diverse correct strategies, consistent with the accuracy drops.

\textbf{Diversity collapse resides in the learned distribution, not the output format}. Narrow two-teacher SFT data reshapes model outputs, and this effect is not reversed by suppressing CoT at inference. This aligns with findings that CoT in post-trained models can function as post-hoc rationalization \citep{lewislim2025cot} and that CoT can be applied selectively~\citep{sprague2025to}. The model has already converged on its answer distribution during training. The Think vs Instruct comparison (\S\ref{sec:cliffs}) is,  therefore, not confounded by the generation format. The diversity difference between lineages reflects data composition. \textit{Practitioners cannot recover diversity by switching Think models to direct-answer mode, the cost is paid at training time}. We note that we measure final-answer diversity, not reasoning-path diversity.

\subsection{Quality-filtered diversity decomposition}
\label{sec:tradeoff}

\begin{wrapfigure}{r}{0.4\columnwidth}
\centering
\vspace{-25pt}
\includegraphics[width=0.4\columnwidth]{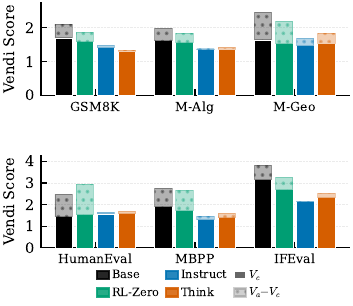}
\caption{Quality filtered Vendi Score on six verifiable tasks.}
\label{fig:qfd_decomposition}
\vspace{8pt}
\includegraphics[width=0.4\columnwidth]{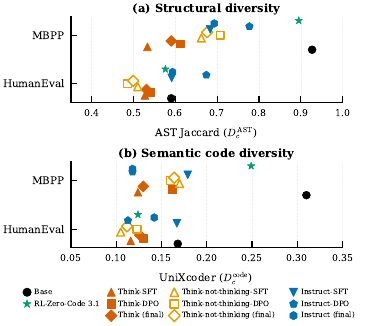}
\caption{Code diversity on correct outputs: AST subtree Jaccard (structural) and UniXcoder (semantic) for HumanEval and MBPP.}
\label{fig:code_div}
\vspace{-25pt}
\end{wrapfigure}

The aggregate diversity reductions combine two effects, elimination of incorrect outputs and genuine narrowing of the correct-answer distribution (Figure~\ref{fig:qfd_decomposition}). We decompose these using $D_a$, $D_c$, $V_a$ and $V_c$ on six verifiable tasks (GSM8K, MATH-Algebra, MATH-Geometry, HumanEval, MBPP, IFEval). All models achieve 94--97\% pass@16 on GSM8K, the underlying capability is broadly present. RL-Zero variants also reach 94--97\% pass@16 on GSM8K despite 49--61\% accuracy@1, confirming the gap is in reliability, not capability. The difference lies in per-attempt reliability (Think 93\% vs.\ Base 56\%), not in whether the knowledge exists.

The proportion of collapse attributable to quality control varies by task (Figure~\ref{fig:qfd_decomposition}; Appendix~\ref{app:qfd}): on IFEval, 83.4\% of the $D_a$ drop persists in $D_c$ (genuine narrowing), while on MBPP 38\% is genuine and on HumanEval less than 10\%. Math reasoning falls between (57--64\% genuine). Code-specific metrics sharpen this picture: among correct HumanEval outputs, Think produces structurally homogeneous solutions (AST Jaccard ${=}0.53$, UniXcoder $D_c{=}0.13$) while Base/RL-Zero's correct outputs are structurally diverse (AST Jaccard ${=}0.89$ on MBPP; Figure~\ref{fig:code_div}). This resolves the tension between \textit{diversity collapse is harmful} and \textit{it is just quality control}~\citep{lake2025overton}: both are right, in task-dependent proportions.

\begin{wrapfigure}{r}{0.4\columnwidth}
\centering
\vspace{-25pt}
\includegraphics[width=0.4\columnwidth]{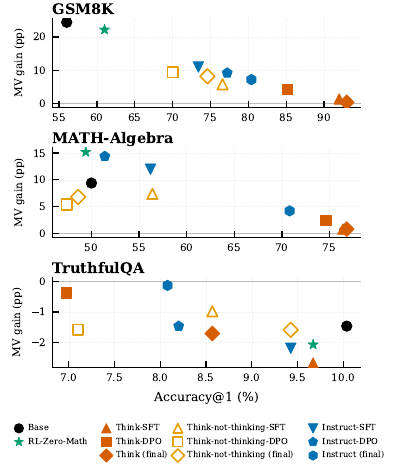}
\vspace{-10pt}
\caption{Accuracy@1 vs.\ majority-voting gain.}
\label{fig:mv_gain}
\vspace{-20pt}
\end{wrapfigure}

Even among correct outputs, a narrowing persists: Base maintains 1.7 effective Vendi modes among its ${\sim}$8.5/16 correct answers, while both Think and Instruct converge to 1.3--1.6 modes among their correct answers (${\sim}$15/16  for GSM8K), while IFEval is higher at 2.1--2.3. In absolute terms, all post-trained models produce near-homogeneous correct outputs, which limits the effectiveness of majority voting~\citep{wang2023selfconsistency}: Think gains just +0.4\% on GSM8K (16 near-identical correct answers provide no independent signal), while Base gains +24\% and RL-Zero +22--26\%. Correct-answer diversity determines how much models benefit from repeated sampling~\citep{snell2025scaling}. On MATH-Algebra, Think-not-thinking and RL-Zero-Math both achieve ${\sim}$49\% accuracy, but RL-Zero-Math has twice the correct-answer diversity and gains +15\% from majority voting compared to +7\% for Think-not-thinking. The pattern holds across math tasks (Figure~\ref{fig:mv_gain}): at matched accuracy, models with more diverse correct outputs consistently extract more benefit from sampling. 

On HumanEval, Instruct surpasses Think at pass@16 (98.2 vs.\ 95.7) despite trailing at pass@1 (81.2 vs.\ 87.7). The collapsed output distribution means additional samples yield identical solutions. On TruthfulQA, the effect is reversed, majority-voting actually \emph{hurts} all models (majority vote@16 $<$ accuracy@1), because the model converges confidently onto the misconception the question was designed to test. When the dominant mode is wrong, diversity collapse amplifies the error. Figure~\ref{fig:mv_gain} visualizes this pattern, high-accuracy models cluster near zero MV gain, while lower-accuracy models with diverse correct outputs benefit substantially. Full quality results are in Appendix~\ref{app:quality}; quality-filtered results in Appendix~\ref{app:qfd}.

\subsection{Cross-cutting patterns}
\label{sec:crosscutting}

The ordering (Base $>$ RL-Zero $>$ Final) holds on average across all 15 tasks, though individual RL-Zero variants exceed Base on tasks aligned with their reward signal (e.g., RL-Zero-IF on IFEval, RL-Zero-Code$^{3.1}$ on HumanEval). A model that is low-diversity on one task tends to be low-diversity on all tasks. Output length does not explain diversity ordering (Appendix~\ref{app:length}).

LLM-as-a-judge evaluation (Figure~\ref{fig:quality}) confirms post-training improves quality across all non-verified task categories. CNN/DM and XSum win rates increase from 26--48\% (Base) to 83--95\% (Think, Instruct), open-ended pairwise win rates exceed 80\% for Think on Alpaca and for both Think and Instruct on PRISM. WildBench scores rise from $-2.0$ (Base) to $6.1$ (Instruct). RL-Zero models are tied with Base on WritingPrompts (50\% win rate), consistent with the absence of creative-writing reward signals. Diversity reductions coexist with clear quality gains.

Among RL-Zero variants, the reward signal type predicts diversity preservation. RL-Zero-IF (instruction-following rewards) retains 99\% of Base diversity on average, while RL-Zero-Code retains only 88\%. On code tasks specifically, RL-Zero-Code retains \emph{less} diversity (90\%) than RL-Zero-General (100\%). Pass/fail execution rewards narrow the solution space more aggressively than general rewards. Mathematical reasoning rewards, which admit diverse solution paths, fall between these extremes. This order (format rewards $>$ math rewards $>$ code rewards) shows that the reward specificity predicts diversity reduction. However, RL-Zero's diversity advantage comes at a steep quality cost, the RL-Zero range is 49.8-61.0\% on GSM8K (vs.\ 93\% Think, 80\% Instruct) and 49\% on IFEval (vs.\ 79\% Think).

\section{Discussion}
\label{sec:discussion}

\textbf{Data composition co-varies with the trajectory, not the floor.}
Think and Instruct share the same three-stage training yet collapse at different stages. The DPO asymmetry (\S\ref{sec:cliffs}) reflects the upstream SFT state more than DPO data differences. Think collapses uniformly across all tasks at SFT, leaving DPO little to remove, while Instruct enters DPO with residual spread that is aggressively narrowed.
Despite these different paths, both lineages converge to 1.3--1.6 Vendi modes among correct answers on most verifiable tasks and ${\sim}$2 modes overall, with IFEval as an outlier at 2.1--2.3. \textit{Data composition} determines \emph{when} and \emph{how sharply} models reach the diversity floor, but not the floor itself. This distinction matters practically, data-level interventions (more teachers, broader sources) can slow the descent but may not raise the final diversity level. Algorithmic changes, switching from reverse to forward KL~\citep{wang2024beyond}, adding entropy constraints~\citep{pan2026qempo}, or removing KL penalties entirely (as in RL-Zero), appear necessary to shift the floor. For SFT data, this suggests that the number of distinct completion sources matters. \textit{Practitioners should avoid single-teacher or dual-teacher distillation when output diversity is valued, and instead draw from multiple models with diverse training}. 

\textbf{Mechanistic interpretation.}
SFT via cross-entropy loss on narrow data performs maximum-likelihood estimation on a low-entropy target distribution. As two teachers from related training lineages produce completions occupying a restricted region of the output space, the model reproduces this narrow mixture. DPO's reverse-KL objective is mode-seeking by construction, its gradient is proportional to the implicit reward gap between chosen and rejected outputs. When the model is already collapsed (Think post-SFT), chosen and rejected responses are both near the mode, yielding small gradients and minimal further compression. When the model retains spread (Instruct post-SFT), DPO aggressively downweights the tails. GRPO \emph{without KL regularization} frees the policy to rediscover modes that SFT and DPO suppressed, provided they receive a positive reward signal.

\textbf{Task-dependent patterns: where diversity loss matters most.}
On \textit{math and reasoning tasks} a significant part of diversity reduction reflects removal of incorrect solution paths, as the narrowing among correct outputs is modest. On \textit{code} tasks, less collapse is genuine narrowing, but it still limits pass@$k$ scaling. \textit{Summarization} shows the largest semantic diversity loss, but this is the cost for large quality gains. \textit{Creative writing and value-pluralism} are the tasks where the observed diversity loss risks imposing a single perspective. The pattern that emerges is a spectrum, from tasks where collapse is largely helpful (code correctness filtering) to tasks where it is actively harmful (value-laden open-ended generation). \textit{Practitioners should assess diversity impact relative to their task characteristics, when selecting post-trained models or applying uniform post-training recipes}.

\textbf{From distributional to representational diversity.}
We capture \emph{distributional} diversity, i.e. statistical spread along lexical, semantic, and logical axes. This is not a sufficient condition for \emph{representational} diversity, the presence of outputs reflecting different perspectives or stances. We detect when a model's output distribution narrows but cannot determine which perspectives are lost. The distinction matters most on value-pluralism tasks. Narrow training data does not just reduce variation, it risks imposing a single perspective on questions where legitimate disagreement exists. A model could maintain high distributional diversity while eliminating viewpoints, or conversely appear collapsed while preserving the stances that matter most. Targeted probes for representational diversity across demographic and cultural dimensions are needed to close this gap.

\section{Conclusion}
\label{sec:conclusion}

We traced output diversity through three parallel post-training lineages of Olmo 3, showing that diversity collapse is shaped by training data composition, not the post-training method alone. The same three-stage recipe (SFT$\to$DPO$\to$RL) produces different collapse trajectories depending on the upstream data: narrow two-teacher distillation drives a steep SFT cliff, while broader multi-source data shifts the sharpest drop to DPO. Suppressing the CoT generation format at inference costs accuracy, but does not recover diversity,  confirming that the collapse resides in the learned weights. Decomposing the diversity loss into quality-control and residual components reveals a task-dependent split. On some tasks nearly all narrowing reflects the removal of errors, on others most of it is genuine homogenization among correct outputs. This directly affects inference scaling and majority voting boosts. For practitioners, our results point to two actionable directions: (1)~broadening the source distribution for SFT data (more teachers, more styles) can mitigate the steepest collapse, and (2)~RL without KL penalties can partially reverse DPO-induced semantic narrowing, though the effect is modest. Future work should investigate reasoning-path diversity (as distinct from final-answer diversity), test data-composition interventions directly, and examine whether the diversity floor we observe can be lowered by changes to the preference-optimization objective.

\section*{Acknowledgments}

We would like to thank Samuel Lewis-Lim for his valuable feedback. CK is supported by the Centre for Doctoral Training in Speech and Language Technologies (SLT) and their Applications funded by UK Research and Innovation grant [grant number EP/S023062/1]. XT and NA are supported by the EPSRC [grant number EP/Y009800/1], through funding from Responsible AI UK (KP0016) as a Keystone project. We acknowledge (1) IT Services at the University of Sheffield for the provision of services for high-performance computing; (2) the use of the University of Oxford Advanced Research Computing (ARC) facility; (3) the EuroHPC Joint Undertaking for awarding this project access to the EuroHPC supercomputer LEONARDO, hosted by CINECA (Italy) and the LEONARDO consortium through an EuroHPC Development Access call; (4) the use of resources provided by the Isambard-AI National AI Research Resource (AIRR). Isambard-AI is operated by the University of Bristol and is funded by the UK Government’s Department for Science, Innovation and Technology (DSIT) via UK Research and Innovation; and the Science and Technology Facilities Council [ST/AIRR/I-A-I/1023].

\bibliographystyle{colm2026_conference}
\bibliography{references}

\clearpage

\appendix

\section{Implementation details}
\label{app:implementation}

We generate outputs using vLLM~\citep{kwon2023efficient} and lighteval~\citep{lighteval}. For each model--task pair, we sample $K{=}16$ outputs per prompt ($N{=}500$ prompts; full dataset for Math-Geometry, IFEval, HumanEval, and TruthfulQA) with a 32{,}768-token generation length. All four diversity metrics (EAD, SBERT, NLI, Vendi Score) operate on the same post-stripping text. Table~\ref{tab:tasks} lists all evaluation tasks with their sample sizes.

\begin{table}[h]
\centering
\label{tab:tasks}
\footnotesize
\resizebox{\columnwidth}{!}{%
\setlength{\tabcolsep}{4pt}
\begin{tabular}{llr@{\qquad}llr@{\qquad}llr}
\toprule
\textbf{Category} & \textbf{Task} & $N$ & \textbf{Category} & \textbf{Task} & $N$ & \textbf{Category} & \textbf{Task} & $N$ \\
\midrule
Summarization & TL;DR & 500 & Reasoning & GSM8K & 500 & Code & HumanEval & 164 \\
      & CNN/DM & 500 &         & MATH-Alg & 500 &      & MBPP & 500 \\
      & XSum & 500   &         & MATH-Geo & 479 &      & CRUXEval & 500 \\
Instruction & Alpaca & 500 &   & TruthfulQA & 817 & Value plur. & PRISM & 500 \\
       & IFEval & 541 & Creative & WrtPrompts & 500 &           & WildBench & 500 \\
\bottomrule
\end{tabular}}
\caption{Evaluation tasks grouped by category.}
\end{table}

\section{Metric definitions}
\label{app:metric_details}

For a given prompt, the model generates $K$ outputs $\{o_1, \ldots, o_K\}$. All metrics are computed per prompt and then averaged over prompts.

\textbf{EAD (lexical diversity)}

Expectation-Adjusted Distinct $n$-grams~\citep{liu-etal-2022-rethinking} counts the number of unique $n$-grams in the output set, normalized by the expected number of unique $n$-grams under a uniform draw from a vocabulary of size $V$. For a total of $T$ $n$-gram tokens with $U$ unique types,
\(
\text{EAD}_n = \frac{U}{V \cdot \left(1 - \left(\frac{V-1}{V}\right)^{T}\right)}\,,
\)
where $V$ is auto-detected from the model's tokenizer vocabulary. The denominator corrects for length bias: longer outputs are expected to contain more unique $n$-grams by chance. We average across $n \in \{1, \ldots, 5\}$ and clip to $[0, 1]$:
\(
D_{\text{EAD}} = \frac{1}{5}\sum_{n=1}^{5}\text{EAD}_n\,.
\)

\textbf{SBERT (semantic diversity)}

We encode each output $o_i$ with \texttt{all-mpnet-base-v2}~\citep{reimers-gurevych-2019-sentence} to obtain L2-normalized embeddings $\mathbf{e}_i$. Semantic diversity is the mean pairwise cosine distance:
\(
D_{\text{SBERT}} = 1 - \frac{2}{K(K-1)} \sum_{i < j} \cos(\mathbf{e}_i, \mathbf{e}_j)\,.
\)
Values near 0 indicate semantic collapse (all outputs map to the same region of embedding space); values near 1 indicate highly dissimilar outputs. For code tasks we additionally report diversity using UniXcoder~\citep{guo2022unixcoder}, a code-aware encoder that captures structural similarity beyond surface tokens.

\textbf{NLI (logical diversity)}

Following~\citet{stasaski-hearst-2022-semantic}, we score output pairs with a natural language inference classifier (\texttt{roberta-large-mnli};~\citealp{liu2019robertarobustlyoptimizedbert}). For each ordered pair $(o_i, o_j)$, the model predicts a probability distribution over \{entailment, neutral, contradiction\}. We compute a directional similarity score as $P(\text{entailment}) - P(\text{contradiction})$, then symmetrize by averaging both orderings:
\(
s_{ij} = \frac{1}{2}\bigl[\bigl(P_{\text{ent}}(o_i \mid o_j) - P_{\text{con}}(o_i \mid o_j)\bigr) + \bigl(P_{\text{ent}}(o_j \mid o_i) - P_{\text{con}}(o_j \mid o_i)\bigr)\bigr]\,.
\)
Since NLI models are trained on single sentences rather than full paragraphs, we align sentences by position across outputs. The diversity score is:
\(
D_{\text{NLI}} = 1 - \frac{2}{K(K-1)}\sum_{i<j} s_{ij}\,.
\)
$D_{\text{NLI}}$ near 0 indicates mutual entailment (collapse), near 1 indicates neutrality, and values above 1 indicate net contradiction (the outputs make mutually inconsistent claims). Code tasks are excluded as NLI is not meaningful for program text.

\textbf{Vendi Score}

The Vendi Score~\citep{friedman2023the} measures the effective number of dissimilar elements via the eigenvalue entropy of a similarity kernel. We reuse the SBERT cosine similarity matrix. Given $K$ outputs with L2-normalized embeddings, we form the Gram matrix $\mathbf{G}$ where $G_{ij} = \cos(\mathbf{e}_i, \mathbf{e}_j)$ and trace-normalize it as $\mathbf{P} = \mathbf{G} / K$. The Vendi Score is
\(
\text{VS} = \exp\!\left(-\sum_i \lambda_i \log \lambda_i\right),
\)
where $\lambda_i$ are the eigenvalues of $\mathbf{P}$. VS${=}1$ when all outputs are identical (rank-1 kernel) and VS${=}K$ when all outputs are orthogonal (full-rank uniform spectrum). Because the Vendi Score shares the SBERT kernel, agreement between VS and $D_{\text{SBERT}}$ is expected rather than independent confirmation; VS adds the interpretable ``effective number of modes'' framing.

\textbf{AST subtree diversity (structural, code only)}

For code-generation tasks (HumanEval, MBPP), we measure structural diversity via the mean pairwise Jaccard distance on AST subtree multisets (subtree height $\le 4$;~\citealp{shypula2025evaluating}). We parse each output into a Python AST, extract all subtrees up to height~4, represent each output as a multiset of subtree hashes, and compute
\(
D_{\text{AST}}(o_i, o_j) = 1 - \frac{|S_i \cap S_j|}{|S_i \cup S_j|}\,,
\)
where $S_i$ is the multiset of subtree hashes for output $o_i$. This metric is reported on correct (executable, test-passing) outputs only, to capture genuine structural variation among working solutions. Unparseable outputs are excluded.

\textbf{LLM-as-a-Judge quality}

For the eight tasks without verifiable answers, we evaluate quality using established LLM-as-judge frameworks with \texttt{gpt-4.1-mini} via the OpenAI Batch API. \emph{Summarization} (TL;DR, CNN/DM, XSum): pairwise win-rate against reference summaries, following~\citet{kirk2024understanding}. \emph{Instruction following and value pluralism} (Alpaca, PRISM): pairwise comparison against Base using MT-Bench prompts~\citep{zheng2023judging}. \emph{Creative writing} (WritingPrompts): pairwise comparison using Arena-Hard creative writing prompts~\citep{li2024arenahard}. \emph{WildBench}: checklist-guided WB-Score~\citep{lin2025wildbench}.
We note that LLM-judge evaluation of creative and value-laden tasks has known limitations~\citep{lu2026rethinking}; we report these results as supplementary context for our diversity findings rather than as primary evidence.

\section{Per-task diversity results}
\label{app:tables}

Tables~\ref{tab:app_sbert}--\ref{tab:app_vendi} report per input diversity for each of the four metrics across all 15 tasks and 16 models (13 standard + 3 Think w/o CoT). Table~\ref{tab:app_sbert} reports SBERT cosine distance, our primary semantic diversity measure. Table~\ref{tab:app_ead} reports Expected Agreement Diversity (EAD), a lexical overlap metric. Table~\ref{tab:app_nli} reports NLI-based diversity, which captures inferential disagreement between output pairs; code tasks are excluded as NLI is not meaningful for program text. Table~\ref{tab:app_vendi} reports Vendi Score, the effective number of distinct semantic modes among the $K{=}16$ outputs.

\input{tables/app_sbert}
\input{tables/app_ead}
\input{tables/app_nli}
\input{tables/app_vendi}

\section{Quality results}
\label{app:quality}

Tables~\ref{tab:app_quality_reasoning}--\ref{tab:app_quality_wildbench} report task performance for all 16 models, across all 15 tasks. Table~\ref{tab:app_quality_reasoning} reports reasoning quality on four tasks (GSM8K, MATH-Algebra, MATH-Geometry, TruthfulQA) with accuracy@1, majority vote@16, and pass@16. Table~\ref{tab:app_quality_code} reports code generation quality (pass@$k$ for $k \in \{1,5,10,16\}$) on HumanEval and MBPP. Table~\ref{tab:app_quality_ifeval} reports IFEval constraint satisfaction with strict and loose accuracy@1, pass@16, and consistency@16. Table~\ref{tab:app_quality_cruxeval} reports CruxEval output-prediction accuracy.

For tasks without verifiable answers, we use LLM-as-judge evaluation with gpt-4.1-mini via the OpenAI Batch API. Table~\ref{tab:app_quality_summarization} reports pairwise win rates against reference summaries following \citet{kirk2024understanding}. Table~\ref{tab:app_quality_openended} reports pairwise win rates against the Base model using the MT-Bench prompt~\citep{zheng2023judging} for Alpaca and PRISM and the Arena-Hard creative writing prompt~\citep{li2024arenahard} for WritingPrompts. Table~\ref{tab:app_quality_wildbench} reports checklist-guided WB-Score~\citep{lin2025wildbench}. We note that LLM-judge evaluation of creative and value-laden tasks has known limitations~\citep{lu2026rethinking}; we report these results as supplementary context for our diversity findings rather than as primary evidence.

\input{tables/app_quality_reasoning}
\input{tables/app_quality_code}
\input{tables/app_quality_ifeval}
\input{tables/app_quality_cruxeval}
\input{tables/app_quality_summarization}
\input{tables/app_quality_openended}
\input{tables/app_quality_wildbench}

\section{Quality-filtered diversity}
\label{app:qfd}

Table~\ref{tab:app_qfd} reports the quality-filtered diversity decomposition defined in \S\ref{sec:metrics} for six verifiable tasks. We label each of $K{=}16$ generations as correct or incorrect (answer matching for math, test execution for code, constraint satisfaction for IFEval), then report accuracy alongside $D_a$ (SBERT on all outputs), $D_c$ (SBERT on correct-only subset, $K_c \geq 2$), and $V_c$ (Vendi Score on correct outputs, interpreted as the effective number of distinct correct answers).

\input{tables/app_qfd}

\section{Code-specific diversity}
\label{app:code_diversity}

Table~\ref{tab:code_diversity} reports quality-filtered code diversity using the domain-specific metrics described in \S\ref{sec:metrics}: UniXcoder SBERT ($D_c^{\text{code}}$, computed on correct outputs only) and AST subtree Jaccard distance ($D_c^{\text{AST}}$, for code-generation tasks). Missing entries (``---'') indicate models with no parseable correct outputs.

\input{tables/app_code_diversity}

\clearpage
\section{Output length analysis}
\label{app:length}

Table~\ref{tab:app_length} reports the mean output word length and mean SBERT diversity per task, averaged across all 13 models. Tasks with high mean diversity (e.g.\ WritingPrompts, HumanEval) span a wide range of output lengths, and tasks with similar lengths (e.g.\ GSM8K at 137 words, TruthfulQA at 142 words) have very different diversity levels (0.128 vs.\ 0.262). Output length does not systematically predict diversity.

\begin{table}[h!]
\centering
\label{tab:app_length}
\footnotesize
\setlength{\tabcolsep}{4pt}
\begin{tabular}{lrr@{\qquad}lrr}
\toprule
\textbf{Task} & \textbf{Len} & \textbf{SBERT} & \textbf{Task} & \textbf{Len} & \textbf{SBERT} \\
\midrule
WildBench & 872 & 0.230 & TL;DR & 283 & 0.270 \\
PRISM & 723 & 0.260 & MATH-Algebra & 227 & 0.110 \\
WritingPrompts & 704 & 0.397 & MBPP & 213 & 0.198 \\
CRUXEval & 619 & 0.183 & HumanEval & 211 & 0.280 \\
MATH-Geometry & 441 & 0.155 & XSum & 158 & 0.292 \\
IFEval & 391 & 0.257 & TruthfulQA & 142 & 0.262 \\
Alpaca & 304 & 0.204 & GSM8K & 137 & 0.128 \\
 & & & CNN/DailyMail & 120 & 0.157 \\
\bottomrule
\end{tabular}
\caption{Mean output word length and SBERT diversity per task, averaged across 13 models.}
\end{table}

\section{Temperature sensitivity}
\label{app:temperature}

Table~\ref{tab:app_temperature} compares Base model diversity at its recommended sampling temperature ($T{=}1.0$, top-$p{=}0.7$) with the matched temperature used throughout this study ($T{=}0.6$, top-$p{=}0.95$). SBERT diversity decreases by 11\% on average, EAD by 18\%, and NLI by only 3\%. These reductions are modest relative to the 62\% SBERT drop from Base to Think-SFT, confirming that the diversity gaps documented in this paper are not attributable to the temperature difference.

\input{tables/app_temperature}

\section{Stage attribution per task}
\label{app:stage_attribution}

Table~\ref{tab:stage_attribution_full} reports the percentage of Base SBERT diversity lost at each post-training stage for all 15 tasks. Think collapses 45--80\% at SFT (most on XSum, least on IFEval), with DPO contributing minimally. Instruct shows the opposite pattern: SFT losses range from 8--73\%, but DPO contributes 2--63\% additional loss. RL-Zero retains 71--105\% of Base diversity across tasks.

\input{tables/stage_attribution_full}

\section{Decontamination}
\label{app:decontamination}

We measure training--evaluation data overlap using $C_{13}$ 13-gram matching~\citep{lambert2025tulu}: for each test instance, we extract all 13-grams (tokenized with spaCy), query an Elasticsearch index of the training data for phrase matches, and report the fraction of test tokens covered by at least one matching 13-gram, averaged over all test instances.
Table~\ref{tab:decontamination} reports results for the four Dolci post-training datasets against all fifteen evaluation benchmarks.
Summarization, creative writing, open-ended QA, and value-pluralism benchmarks show negligible overlap (${\le}\,1.6\%$).
HumanEval, CRUXEval, IFEval, MATH, and WildBench show elevated overlap (7--30\%), traceable to shared upstream sources: the Dolci SFT mixes include OpenThoughts3 \citep{guha2025openthoughtsdatarecipesreasoning}, whose math questions derive from OpenMathInstruct-2 \citep{toshniwal2024openmath2}, itself built on the MATH training set, large-scale Python corpora, Dolci-Think-Python \citep{olmo2025olmo3}, Nemotron \citep{nvidia_nemotron_nano_v3_2025} code split, and WildChat conversations \citep{zhao2024wildchat}.

\input{tables/decontamination}

\end{document}

%% file: tables/stage_attribution.tex
\begin{wraptable}{r}{0.4\columnwidth}
\vspace{-1pt}
\centering
\label{tab:stage_attribution}
\scriptsize
\setlength{\tabcolsep}{3pt}
\begin{tabular}{@{}l ccc c@{}}
\toprule
& SFT & DPO & RL & Retained \\
\midrule
Think & $-$62 & $-$4 & $+$4 & 38\% \\
Instruct & $-$38 & $-$23 & $-$5 & 34\% \\
\midrule
RL-Zero & \multicolumn{3}{c}{(single)} & 93\% \\
\bottomrule
\end{tabular}
\caption{Stage-wise SBERT loss (\% of Base, 15-task average).}
\label{tab:stages}
\vspace{-8pt}
\end{wraptable}

%% file: tables/app_sbert.tex
\begin{table*}[htbp]
\centering
\resizebox{\textwidth}{!}{%
\begin{tabular}{lcccccccc}
\toprule
 & \multicolumn{3}{c}{\textit{Summarization}} & \multicolumn{2}{c}{\textit{Instruction F.}} & \multicolumn{1}{c}{\textit{Creative Wr.}} & \multicolumn{2}{c}{\textit{Value Pluralism}} \\
\cmidrule(lr){2-4}\cmidrule(lr){5-6}\cmidrule(lr){7-7}\cmidrule(lr){8-9}
 & TL;DR & CNN/DM & XSum & Alpaca & IFEval & WritingPrompts & PRISM & WildBench \\
\midrule
Base & 0.353 & 0.279 & 0.451 & 0.319 & 0.349 & 0.540 & 0.408 & 0.335 \\
\midrule
Instruct-SFT & 0.268 & 0.223 & 0.282 & 0.170 & 0.172 & 0.276 & 0.141 & 0.129 \\
Instruct-DPO & 0.202 & 0.075 & 0.083 & 0.120 & 0.154 & 0.225 & 0.096 & 0.122 \\
Instruct (final) & 0.207 & 0.072 & 0.081 & 0.113 & 0.154 & 0.202 & 0.090 & 0.118 \\
\midrule
Think-SFT & 0.168 & 0.083 & 0.090 & 0.141 & 0.191 & 0.240 & 0.100 & 0.160 \\
Think-DPO & 0.159 & 0.059 & 0.064 & 0.118 & 0.165 & 0.205 & 0.089 & 0.154 \\
Think (final) & 0.161 & 0.091 & 0.092 & 0.146 & 0.196 & 0.199 & 0.091 & 0.173 \\
\midrule
Think-SFT w/o CoT & 0.293 & 0.249 & 0.202 & 0.137 & 0.196 & 0.266 & 0.114 & 0.191 \\
Think-DPO w/o CoT & 0.344 & 0.176 & 0.130 & 0.104 & 0.157 & 0.223 & 0.100 & 0.156 \\
Think w/o CoT & 0.323 & 0.220 & 0.167 & 0.161 & 0.221 & 0.245 & 0.102 & 0.181 \\
\midrule
RL-Zero-Math & 0.336 & 0.201 & 0.436 & 0.309 & 0.318 & 0.543 & 0.393 & 0.313 \\
RL-Zero-Code & 0.327 & 0.193 & 0.422 & 0.178 & 0.287 & 0.533 & 0.367 & 0.262 \\
RL-Zero-IF & 0.333 & 0.210 & 0.429 & 0.176 & 0.397 & 0.546 & 0.400 & 0.300 \\
RL-Zero-General & 0.309 & 0.184 & 0.404 & 0.155 & 0.284 & 0.523 & 0.372 & 0.279 \\
RL-Zero-Math$^{3.1}$ & 0.330 & 0.200 & 0.432 & 0.319 & 0.324 & 0.546 & 0.398 & 0.316 \\
RL-Zero-Code$^{3.1}$ & 0.328 & 0.196 & 0.430 & 0.314 & 0.325 & 0.539 & 0.394 & 0.315 \\
\bottomrule
\end{tabular}}
\vspace{1pt}
\resizebox{\textwidth}{!}{%
\begin{tabular}{lccccccc}
\toprule
 & \multicolumn{4}{c}{\textit{Reasoning}} & \multicolumn{3}{c}{\textit{Code}} \\
\cmidrule(lr){2-5}\cmidrule(lr){6-8}
 & GSM8K & MATH-Alg & MATH-Geo & TruthfulQA & HumanEval & MBPP & CRUXEval \\
\midrule
Base & 0.172 & 0.146 & 0.198 & 0.353 & 0.411 & 0.291 & 0.239 \\
\midrule
Instruct-SFT & 0.105 & 0.132 & 0.179 & 0.327 & 0.112 & 0.111 & 0.218 \\
Instruct-DPO & 0.141 & 0.071 & 0.096 & 0.158 & 0.095 & 0.073 & 0.068 \\
Instruct (final) & 0.078 & 0.057 & 0.101 & 0.115 & 0.093 & 0.069 & 0.062 \\
\midrule
Think-SFT & 0.061 & 0.054 & 0.107 & 0.119 & 0.109 & 0.081 & 0.095 \\
Think-DPO & 0.052 & 0.061 & 0.114 & 0.074 & 0.081 & 0.084 & 0.076 \\
Think (final) & 0.051 & 0.062 & 0.122 & 0.075 & 0.117 & 0.089 & 0.090 \\
\midrule
Think-SFT w/o CoT & 0.057 & 0.066 & 0.098 & 0.106 & 0.055 & 0.084 & 0.084 \\
Think-DPO w/o CoT & 0.045 & 0.058 & 0.077 & 0.085 & 0.062 & 0.083 & 0.064 \\
Think w/o CoT & 0.052 & 0.064 & 0.089 & 0.089 & 0.060 & 0.083 & 0.071 \\
\midrule
RL-Zero-Math & 0.154 & 0.144 & 0.181 & 0.352 & 0.421 & 0.274 & 0.222 \\
RL-Zero-Code & 0.156 & 0.144 & 0.183 & 0.348 & 0.464 & 0.238 & 0.149 \\
RL-Zero-IF & 0.177 & 0.143 & 0.199 & 0.357 & 0.336 & 0.297 & 0.491 \\
RL-Zero-General & 0.133 & 0.124 & 0.166 & 0.326 & 0.468 & 0.272 & 0.198 \\
RL-Zero-Math$^{3.1}$ & 0.183 & 0.140 & 0.183 & 0.358 & 0.460 & 0.292 & 0.207 \\
RL-Zero-Code$^{3.1}$ & 0.173 & 0.139 & 0.178 & 0.349 & 0.439 & 0.261 & 0.209 \\
\bottomrule
\end{tabular}}
\caption{Per-input \textbf{SBERT} diversity (\texttt{all-mpnet-base-v2}).}
\label{tab:app_sbert}
\end{table*}

%% file: tables/app_ead.tex
\begin{table*}[htbp]
\centering
\resizebox{\textwidth}{!}{%
\begin{tabular}{lcccccccc}
\toprule
 & \multicolumn{3}{c}{\textit{Summarization}} & \multicolumn{2}{c}{\textit{Instruction F.}} & \multicolumn{1}{c}{\textit{Creative Wr.}} & \multicolumn{2}{c}{\textit{Value Pluralism}} \\
\cmidrule(lr){2-4}\cmidrule(lr){5-6}\cmidrule(lr){7-7}\cmidrule(lr){8-9}
 & TL;DR & CNN/DM & XSum & Alpaca & IFEval & WritingPrompts & PRISM & WildBench \\
\midrule
Base & 0.37 & 0.37 & 0.67 & 0.51 & 0.44 & 0.23 & 0.24 & 0.30 \\
\midrule
Instruct-SFT & 0.69 & 0.43 & 0.58 & 0.57 & 0.62 & 0.72 & 0.68 & 0.68 \\
Instruct-DPO & 0.71 & 0.53 & 0.51 & 0.56 & 0.71 & 0.80 & 0.72 & 0.76 \\
Instruct (final) & 0.68 & 0.50 & 0.48 & 0.52 & 0.67 & 0.79 & 0.70 & 0.74 \\
\midrule
Think-SFT & 0.76 & 0.59 & 0.58 & 0.61 & 0.58 & 0.73 & 0.72 & 0.63 \\
Think-DPO & 0.79 & 0.62 & 0.63 & 0.69 & 0.74 & 0.83 & 0.76 & 0.75 \\
Think (final) & 0.78 & 0.59 & 0.59 & 0.65 & 0.68 & 0.80 & 0.74 & 0.71 \\
\midrule
Think-SFT w/o CoT & 0.44 & 0.42 & 0.45 & 0.65 & 0.61 & 0.75 & 0.71 & 0.59 \\
Think-DPO w/o CoT & 0.70 & 0.56 & 0.61 & 0.67 & 0.72 & 0.81 & 0.73 & 0.69 \\
Think w/o CoT & 0.56 & 0.46 & 0.49 & 0.63 & 0.64 & 0.77 & 0.70 & 0.64 \\
\midrule
RL-Zero-Math & 0.41 & 0.38 & 0.64 & 0.49 & 0.55 & 0.33 & 0.36 & 0.44 \\
RL-Zero-Code & 0.47 & 0.39 & 0.66 & 0.58 & 0.60 & 0.40 & 0.43 & 0.53 \\
RL-Zero-IF & 0.46 & 0.41 & 0.66 & 0.47 & 0.55 & 0.33 & 0.39 & 0.50 \\
RL-Zero-General & 0.45 & 0.38 & 0.65 & 0.62 & 0.62 & 0.36 & 0.44 & 0.51 \\
RL-Zero-Math$^{3.1}$ & 0.34 & 0.36 & 0.62 & 0.53 & 0.53 & 0.30 & 0.35 & 0.43 \\
RL-Zero-Code$^{3.1}$ & 0.35 & 0.36 & 0.61 & 0.56 & 0.54 & 0.29 & 0.35 & 0.45 \\
\bottomrule
\end{tabular}}
\vspace{1pt}
\resizebox{\textwidth}{!}{%
\begin{tabular}{lccccccc}
\toprule
 & \multicolumn{4}{c}{\textit{Reasoning}} & \multicolumn{3}{c}{\textit{Code}} \\
\cmidrule(lr){2-5}\cmidrule(lr){6-8}
 & GSM8K & MATH-Alg & MATH-Geo & TruthfulQA & HumanEval & MBPP & CRUXEval \\
\midrule
Base & 0.45 & 0.45 & 0.40 & 0.46 & 0.57 & 0.59 & 0.31 \\
\midrule
Instruct-SFT & 0.38 & 0.45 & 0.51 & 0.57 & 0.48 & 0.51 & 0.57 \\
Instruct-DPO & 0.47 & 0.44 & 0.56 & 0.65 & 0.57 & 0.57 & 0.58 \\
Instruct (final) & 0.36 & 0.39 & 0.52 & 0.64 & 0.55 & 0.54 & 0.55 \\
\midrule
Think-SFT & 0.36 & 0.30 & 0.41 & 0.62 & 0.43 & 0.43 & 0.48 \\
Think-DPO & 0.36 & 0.32 & 0.46 & 0.64 & 0.42 & 0.48 & 0.50 \\
Think (final) & 0.32 & 0.32 & 0.45 & 0.61 & 0.46 & 0.46 & 0.50 \\
\midrule
Think-SFT w/o CoT & 0.41 & 0.43 & 0.50 & 0.69 & 0.40 & 0.54 & 0.52 \\
Think-DPO w/o CoT & 0.37 & 0.41 & 0.49 & 0.72 & 0.46 & 0.58 & 0.51 \\
Think w/o CoT & 0.40 & 0.43 & 0.49 & 0.67 & 0.45 & 0.55 & 0.50 \\
\midrule
RL-Zero-Math & 0.41 & 0.49 & 0.49 & 0.57 & 0.59 & 0.61 & 0.41 \\
RL-Zero-Code & 0.47 & 0.49 & 0.51 & 0.58 & 0.56 & 0.60 & 0.41 \\
RL-Zero-IF & 0.46 & 0.43 & 0.50 & 0.57 & 0.61 & 0.64 & 0.55 \\
RL-Zero-General & 0.45 & 0.45 & 0.47 & 0.52 & 0.57 & 0.59 & 0.45 \\
RL-Zero-Math$^{3.1}$ & 0.41 & 0.46 & 0.48 & 0.54 & 0.57 & 0.60 & 0.40 \\
RL-Zero-Code$^{3.1}$ & 0.43 & 0.47 & 0.49 & 0.54 & 0.56 & 0.59 & 0.43 \\
\bottomrule
\end{tabular}}
\caption{Per-input \textbf{EAD} diversity.}
\label{tab:app_ead}
\end{table*}

%% file: tables/app_nli.tex
\begin{table*}[htbp]
\centering
\footnotesize
\setlength{\tabcolsep}{3pt}
\resizebox{\textwidth}{!}{%
\begin{tabular}{lcccccccc}
\toprule
 & \multicolumn{3}{c}{\textit{Summarization}} & \multicolumn{2}{c}{\textit{Instruction.\ F.}} & \multicolumn{1}{c}{\textit{Creative Wr.}} & \multicolumn{2}{c}{\textit{Value Pluralism}} \\
\cmidrule(lr){2-4}\cmidrule(lr){5-6}\cmidrule(lr){7-7}\cmidrule(lr){8-9}
 & TL;DR & CNN/DM & XSum & Alpaca & IFEval & WritingPrompts & PRISM & WildBench \\
\midrule
Base & 0.95 & 1.04 & 1.09 & 0.68 & 1.05 & 1.16 & 1.09 & 1.06 \\
\midrule
Instruct-SFT & 0.90 & 0.71 & 0.99 & 0.78 & 0.97 & 1.02 & 0.93 & 1.05 \\
Instruct-DPO & 0.86 & 0.84 & 0.77 & 0.77 & 0.98 & 1.05 & 0.97 & 1.06 \\
Instruct (final) & 0.84 & 0.79 & 0.72 & 0.73 & 0.93 & 1.02 & 0.95 & 1.05 \\
\midrule
Think-SFT & 1.02 & 0.93 & 0.92 & 0.89 & 1.01 & 1.13 & 1.04 & 1.09 \\
Think-DPO & 1.06 & 0.93 & 0.93 & 0.93 & 1.06 & 1.18 & 1.07 & 1.12 \\
Think (final) & 1.03 & 0.90 & 0.89 & 0.85 & 1.00 & 1.12 & 1.04 & 1.09 \\
\midrule
Think-SFT w/o CoT & 1.04 & 0.98 & 0.99 & 0.96 & 1.01 & 1.12 & 1.04 & 1.10 \\
Think-DPO w/o CoT & 0.98 & 0.96 & 0.97 & 0.99 & 1.06 & 1.18 & 1.08 & 1.10 \\
Think w/o CoT & 1.00 & 0.98 & 0.97 & 0.91 & 1.00 & 1.09 & 1.02 & 1.09 \\
\midrule
RL-Zero-Math & 0.92 & 0.90 & 1.05 & 0.69 & 1.05 & 1.16 & 1.09 & 1.08 \\
RL-Zero-Code & 0.90 & 0.89 & 1.04 & 0.97 & 1.05 & 1.14 & 1.07 & 1.08 \\
RL-Zero-IF & 0.89 & 0.85 & 1.04 & 0.68 & 0.89 & 1.15 & 1.06 & 1.01 \\
RL-Zero-General & 0.89 & 0.89 & 1.04 & 0.85 & 1.02 & 1.14 & 1.06 & 1.06 \\
RL-Zero-Math$^{3.1}$ & 0.92 & 0.90 & 1.05 & 0.69 & 1.05 & 1.15 & 1.08 & 1.07 \\
RL-Zero-Code$^{3.1}$ & 0.91 & 0.89 & 1.05 & 0.74 & 1.06 & 1.15 & 1.08 & 1.07 \\
\bottomrule
\end{tabular}}
\vspace{4pt}
\resizebox{0.7\textwidth}{!}{%
\begin{tabular}{lcccc}
\toprule
 & \multicolumn{4}{c}{\textit{Reasoning}} \\
\cmidrule(lr){2-5}
 & GSM8K & MATH-Alg & MATH-Geo & TruthfulQA \\
\midrule
Base & 1.08 & 1.00 & 1.13 & 0.97 \\
\midrule
Instruct-SFT & 0.77 & 1.01 & 1.10 & 0.88 \\
Instruct-DPO & 0.77 & 0.76 & 0.88 & 0.91 \\
Instruct (final) & 0.73 & 0.76 & 0.89 & 0.90 \\
\midrule
Think-SFT & 0.77 & 0.72 & 0.85 & 0.98 \\
Think-DPO & 0.73 & 0.72 & 0.86 & 0.98 \\
Think (final) & 0.70 & 0.73 & 0.86 & 0.99 \\
\midrule
Think-SFT w/o CoT & 0.90 & 0.87 & 1.00 & 1.03 \\
Think-DPO w/o CoT & 0.81 & 0.90 & 1.02 & 1.05 \\
Think w/o CoT & 0.87 & 0.91 & 1.03 & 1.00 \\
\midrule
RL-Zero-Math & 1.05 & 0.99 & 1.09 & 0.97 \\
RL-Zero-Code & 1.05 & 0.98 & 1.09 & 0.96 \\
RL-Zero-IF & 1.01 & 0.96 & 1.10 & 0.95 \\
RL-Zero-General & 1.02 & 0.95 & 1.08 & 0.94 \\
RL-Zero-Math$^{3.1}$ & 1.06 & 0.98 & 1.10 & 0.97 \\
RL-Zero-Code$^{3.1}$ & 1.05 & 0.98 & 1.09 & 0.97 \\
\bottomrule
\end{tabular}}
\caption{Per-input \textbf{NLI} diversity. Code tasks excluded.}
\label{tab:app_nli}
\end{table*}

%% file: tables/app_vendi.tex
\begin{table*}[htbp]
\centering
\footnotesize
\setlength{\tabcolsep}{3pt}
\resizebox{\textwidth}{!}{%
\begin{tabular}{lcccccccc}
\toprule
 & \multicolumn{3}{c}{\textit{Summarization}} & \multicolumn{2}{c}{\textit{Instruction F.}} & \multicolumn{1}{c}{\textit{Creative Wr.}} & \multicolumn{2}{c}{\textit{Value Pluralism}} \\
\cmidrule(lr){2-4}\cmidrule(lr){5-6}\cmidrule(lr){7-7}\cmidrule(lr){8-9}
 & TL;DR & CNN/DM & XSum & Alpaca & IFEval & WritingPrompts & PRISM & WildBench \\
\midrule
Base & 4.2 & 3.2 & 5.2 & 2.2 & 3.8 & 6.9 & 4.6 & 3.5 \\
\midrule
Instruct-SFT & 3.0 & 2.4 & 3.2 & 2.1 & 2.3 & 3.2 & 2.0 & 1.9 \\
Instruct-DPO & 2.4 & 1.5 & 1.6 & 1.8 & 2.1 & 2.8 & 1.7 & 1.9 \\
Instruct (final) & 2.5 & 1.5 & 1.5 & 1.7 & 2.2 & 2.6 & 1.6 & 1.8 \\
\midrule
Think-SFT & 2.2 & 1.6 & 1.6 & 2.0 & 2.4 & 2.9 & 1.7 & 2.2 \\
Think-DPO & 2.2 & 1.4 & 1.4 & 1.9 & 2.3 & 2.6 & 1.6 & 2.1 \\
Think (final) & 2.2 & 1.6 & 1.6 & 2.0 & 2.5 & 2.6 & 1.6 & 2.3 \\
\midrule
Think-SFT w/o CoT & 3.0 & 2.5 & 2.2 & 2.0 & 2.4 & 3.1 & 1.8 & 2.3 \\
Think-DPO w/o CoT & 2.8 & 2.0 & 1.8 & 1.7 & 2.1 & 2.7 & 1.6 & 2.0 \\
Think w/o CoT & 3.0 & 2.3 & 2.0 & 2.0 & 2.4 & 2.8 & 1.7 & 2.2 \\
\midrule
RL-Zero-Math & 3.9 & 2.4 & 4.9 & 2.3 & 3.5 & 7.0 & 4.5 & 3.3 \\
RL-Zero-Code & 3.8 & 2.3 & 4.7 & 2.1 & 3.2 & 6.8 & 4.2 & 2.9 \\
RL-Zero-IF & 3.8 & 2.4 & 4.8 & 2.0 & 4.4 & 7.0 & 4.5 & 3.1 \\
RL-Zero-General & 3.6 & 2.3 & 4.5 & 2.0 & 3.2 & 6.7 & 4.2 & 3.0 \\
RL-Zero-Math$^{3.1}$ & 3.8 & 2.4 & 4.8 & 2.2 & 3.6 & 7.0 & 4.6 & 3.3 \\
RL-Zero-Code$^{3.1}$ & 3.8 & 2.4 & 4.8 & 2.3 & 3.5 & 6.9 & 4.5 & 3.3 \\
\bottomrule
\end{tabular}}
\vspace{4pt}
\resizebox{\textwidth}{!}{%
\begin{tabular}{lccccccc}
\toprule
 & \multicolumn{4}{c}{\textit{Reasoning}} & \multicolumn{3}{c}{\textit{Code}} \\
\cmidrule(lr){2-5}\cmidrule(lr){6-8}
 & GSM8K & MATH-Alg & MATH-Geo & TruthfulQA & HumanEval & MBPP & CRUXEval \\
\midrule
Base & 2.1 & 2.0 & 2.4 & 3.8 & 2.5 & 2.8 & 2.5 \\
\midrule
Instruct-SFT & 1.7 & 1.9 & 2.3 & 3.4 & 1.7 & 1.7 & 2.2 \\
Instruct-DPO & 1.9 & 1.5 & 1.6 & 2.0 & 1.7 & 1.5 & 1.5 \\
Instruct (final) & 1.5 & 1.4 & 1.7 & 1.7 & 1.6 & 1.5 & 1.4 \\
\midrule
Think-SFT & 1.4 & 1.4 & 1.7 & 1.8 & 1.6 & 1.5 & 1.6 \\
Think-DPO & 1.3 & 1.4 & 1.8 & 1.5 & 1.5 & 1.6 & 1.5 \\
Think (final) & 1.3 & 1.4 & 1.8 & 1.5 & 1.7 & 1.6 & 1.6 \\
\midrule
Think-SFT w/o CoT & 1.4 & 1.4 & 1.6 & 1.7 & 1.4 & 1.6 & 1.6 \\
Think-DPO w/o CoT & 1.3 & 1.4 & 1.5 & 1.6 & 1.4 & 1.5 & 1.4 \\
Think w/o CoT & 1.3 & 1.4 & 1.6 & 1.6 & 1.4 & 1.5 & 1.5 \\
\midrule
RL-Zero-Math & 2.0 & 2.0 & 2.3 & 3.8 & 2.6 & 2.7 & 2.3 \\
RL-Zero-Code & 2.0 & 2.0 & 2.3 & 3.7 & 3.0 & 2.5 & 1.9 \\
RL-Zero-IF & 2.1 & 1.9 & 2.4 & 3.8 & 2.0 & 2.7 & 3.9 \\
RL-Zero-General & 1.9 & 1.8 & 2.2 & 3.5 & 2.9 & 2.6 & 2.2 \\
RL-Zero-Math$^{3.1}$ & 2.2 & 1.9 & 2.3 & 3.9 & 2.9 & 2.8 & 2.1 \\
RL-Zero-Code$^{3.1}$ & 2.1 & 1.9 & 2.2 & 3.8 & 2.7 & 2.6 & 2.1 \\
\bottomrule
\end{tabular}}
\caption{Per-input \textbf{Vendi Score} diversity.}
\label{tab:app_vendi}
\end{table*}

%% file: tables/app_quality_reasoning.tex
\begin{table*}[htbp]
\centering
\footnotesize
\setlength{\tabcolsep}{3pt}
\resizebox{\textwidth}{!}{%
\begin{tabular}{l ccc ccc ccc ccc}
\toprule
 & \multicolumn{3}{c}{GSM8K} & \multicolumn{3}{c}{MATH-Algebra} & \multicolumn{3}{c}{MATH-Geometry} & \multicolumn{3}{c}{TruthfulQA} \\
\cmidrule(lr){2-4}\cmidrule(lr){5-7}\cmidrule(lr){8-10}\cmidrule(lr){11-13}
 & acc & mv & pass & acc & mv & pass & acc & mv & pass & acc & mv & pass \\
\midrule
Base & 56.0 & 80.4 & 94.8 & 50.0 & 59.4 & 75.6 & 20.5 & 24.6 & 50.3 & 10.0 & 8.6 & 28.5 \\
\midrule
Instruct-SFT & 73.4 & 84.4 & 95.4 & 56.2 & 68.2 & \textbf{83.2} & 26.5 & 36.7 & 61.0 & 9.4 & 7.2 & 24.0 \\
Instruct-DPO & 77.2 & 86.4 & 96.2 & 51.4 & 65.8 & 81.6 & 23.0 & 35.7 & 54.9 & 8.2 & 6.7 & 20.8 \\
Instruct (final) & 80.4 & 87.6 & 95.2 & 70.8 & 75.0 & 81.2 & 42.6 & 54.3 & \textbf{63.3} & 8.1 & 8.0 & 19.6 \\
\midrule
Think-SFT & 92.0 & \textbf{93.4} & \textbf{97.0} & 76.4 & 77.2 & 78.8 & 50.5 & 54.7 & 59.5 & 9.7 & 7.0 & 21.5 \\
Think-DPO & 85.2 & 89.4 & 95.6 & 74.6 & 77.0 & 78.2 & 50.5 & 54.5 & 61.6 & 7.0 & 6.6 & 13.8 \\
Think (final) & \textbf{93.0} & \textbf{93.4} & 96.4 & \textbf{76.8} & \textbf{77.6} & 78.8 & \textbf{51.1} & \textbf{55.3} & 59.7 & 8.6 & 6.9 & 19.3 \\
\midrule
Think-SFT w/o CoT & 76.6 & 82.4 & 94.6 & 56.4 & 63.8 & 75.0 & 27.6 & 29.9 & 46.8 & 8.6 & 7.6 & 16.3 \\
Think-DPO w/o CoT & 70.0 & 79.4 & 94.6 & 47.4 & 52.8 & 67.8 & 19.6 & 23.2 & 37.2 & 7.1 & 5.5 & 12.1 \\
Think w/o CoT & 74.6 & 82.8 & 94.0 & 48.6 & 55.4 & 68.2 & 19.6 & 25.5 & 38.8 & 9.4 & 7.8 & 16.8 \\
\midrule
RL-Zero-Math & 61.0 & 83.2 & 95.8 & 49.4 & 64.6 & 79.2 & 22.8 & 27.3 & 57.6 & 9.7 & 7.6 & 29.4 \\
RL-Zero-Code & 58.2 & 83.8 & 96.4 & 51.2 & 63.4 & 80.8 & 23.0 & 28.2 & 57.8 & \textbf{12.2} & 8.2 & \textbf{30.6} \\
RL-Zero-IF & 49.8 & 75.0 & 94.2 & 48.2 & 60.8 & 77.4 & 21.3 & 25.5 & 51.6 & 10.2 & \textbf{9.3} & 28.8 \\
RL-Zero-General & 61.0 & 82.8 & \textbf{97.0} & 54.0 & 64.8 & 79.6 & 24.4 & 29.6 & 57.2 & 10.9 & 7.7 & 28.8 \\
RL-Zero-Math$^{3.1}$ & 55.2 & 80.6 & 96.2 & 53.6 & 66.0 & 78.8 & 20.9 & 28.4 & 55.7 & 10.6 & 8.4 & 28.2 \\
RL-Zero-Code$^{3.1}$ & 59.8 & 81.8 & 95.2 & 52.4 & 62.8 & 80.4 & 22.1 & 27.1 & 56.8 & 10.2 & 8.2 & 27.5 \\
\bottomrule
\end{tabular}}
\caption{\textbf{Reasoning} quality (\%). acc: first correct. mv: majority vote. pass: any of $K{=}16$ correct.}
\label{tab:app_quality_reasoning}
\end{table*}

%% file: tables/app_quality_code.tex
\begin{table*}[htbp]
\centering
\footnotesize
\setlength{\tabcolsep}{3pt}
\resizebox{0.8\textwidth}{!}{%
\begin{tabular}{l cccc cccc}
\toprule
 & \multicolumn{4}{c}{HumanEval} & \multicolumn{4}{c}{MBPP} \\
\cmidrule(lr){2-5}\cmidrule(lr){6-9}
 & @1 & @5 & @10 & @16 & @1 & @5 & @10 & @16 \\
\midrule
Base & 1.6 & 6.6 & 10.8 & 14.0 & 23.9 & 45.0 & 50.9 & 54.0 \\
\midrule
Instruct-SFT & 63.4 & 88.1 & 93.6 & 96.3 & 32.3 & 47.5 & 52.1 & 54.8 \\
Instruct-DPO & 73.3 & 93.6 & 96.4 & 97.0 & 32.9 & 47.9 & 51.8 & 53.6 \\
Instruct (final) & 81.2 & \textbf{96.2} & \textbf{97.7} & \textbf{98.2} & 37.8 & 48.9 & 51.7 & 53.2 \\
\midrule
Think-SFT & 86.7 & 94.9 & 95.6 & 95.7 & 41.0 & 50.1 & 52.3 & 53.6 \\
Think-DPO & 86.5 & 94.5 & 95.0 & 95.1 & 40.6 & 49.7 & 51.9 & 52.8 \\
Think (final) & \textbf{87.7} & 95.0 & 95.6 & 95.7 & \textbf{44.1} & \textbf{53.7} & \textbf{56.1} & \textbf{58.0} \\
\midrule
Think-SFT w/o CoT & 49.4 & 76.3 & 81.8 & 84.1 & 24.0 & 43.2 & 48.6 & 51.4 \\
Think-DPO w/o CoT & 56.5 & 78.4 & 82.4 & 84.8 & 26.2 & 42.9 & 47.2 & 49.4 \\
Think w/o CoT & 55.6 & 77.6 & 82.0 & 84.1 & 23.9 & 42.1 & 47.7 & 50.8 \\
\midrule
RL-Zero-Math & 2.4 & 10.3 & 17.1 & 23.2 & 24.5 & 45.6 & 51.7 & 55.0 \\
RL-Zero-Code & 2.7 & 11.2 & 19.1 & 26.8 & 24.8 & 44.8 & 50.2 & 53.2 \\
RL-Zero-IF & 1.1 & 5.1 & 9.4 & 13.4 & 24.6 & 45.0 & 51.8 & 56.0 \\
RL-Zero-General & 2.5 & 11.1 & 19.7 & 28.0 & 24.9 & 44.9 & 51.0 & 54.6 \\
RL-Zero-Math$^{3.1}$ & 2.1 & 9.0 & 15.5 & 21.3 & 24.1 & 44.7 & 50.2 & 53.0 \\
RL-Zero-Code$^{3.1}$ & 66.5 & 83.9 & 87.9 & 89.6 & 25.4 & 45.4 & 50.6 & 53.2 \\
\bottomrule
\end{tabular}}
\caption{\textbf{Code} quality (pass@$k$, \%).}
\label{tab:app_quality_code}
\end{table*}

%% file: tables/app_quality_ifeval.tex
\begin{table}[htbp]
\centering
\footnotesize
\setlength{\tabcolsep}{4pt}
\begin{tabular}{l cccc}
\toprule
 & strict@1 & loose@1 & pass@16 & consist \\
\midrule
Base & 44.7 & 58.4 & 74.1 & 46.5 \\
\midrule
Instruct-SFT & 78.7 & 85.9 & 90.6 & 79.3 \\
Instruct-DPO & 78.7 & 85.3 & 89.6 & 79.3 \\
Instruct (final) & \textbf{82.1} & \textbf{87.9} & 89.3 & \textbf{81.8} \\
\midrule
Think-SFT & 78.0 & 84.9 & 90.9 & 77.2 \\
Think-DPO & 74.9 & 81.4 & 86.7 & 73.7 \\
Think (final) & 78.7 & 85.2 & \textbf{91.7} & 79.5 \\
\midrule
Think-SFT w/o CoT & 70.2 & 79.3 & 89.5 & 71.5 \\
Think-DPO w/o CoT & 66.7 & 75.4 & 82.4 & 66.5 \\
Think w/o CoT & 71.0 & 79.6 & 88.0 & 70.7 \\
\midrule
RL-Zero-Math & 47.7 & 59.8 & 72.6 & 47.0 \\
RL-Zero-Code & 47.0 & 60.1 & 71.5 & 46.6 \\
RL-Zero-IF & 59.7 & 70.5 & 75.0 & 61.0 \\
RL-Zero-General & 46.6 & 59.6 & 72.3 & 48.0 \\
RL-Zero-Math$^{3.1}$ & 48.6 & 60.7 & 72.5 & 45.9 \\
RL-Zero-Code$^{3.1}$ & 46.4 & 58.5 & 73.6 & 46.7 \\
\bottomrule
\end{tabular}
\caption{\textbf{IFEval} constraint satisfaction (\%).}
\label{tab:app_quality_ifeval}
\end{table}

%% file: tables/app_quality_cruxeval.tex
\begin{table}[htbp]
\centering
\footnotesize
\setlength{\tabcolsep}{5pt}
\begin{tabular}{lccc}
\toprule
 & Acc@1 & MV@16 & Pass@16 \\
\midrule
Base & 16.4 & 36.0 & 61.0 \\
\midrule
Instruct-SFT & 32.4 & 43.5 & 73.6 \\
Instruct-DPO & \textbf{32.9} & 40.0 & \textbf{84.5} \\
Instruct (final) & 18.0 & 21.0 & 76.2 \\
\midrule
Think-SFT & 19.2 & 28.2 & 74.5 \\
Think-DPO & 17.4 & 26.5 & 65.1 \\
Think (final) & 15.8 & 29.4 & 65.5 \\
\midrule
Think-SFT w/o CoT & 26.3 & \textbf{47.5} & 74.2 \\
Think-DPO w/o CoT & 27.0 & 44.2 & 70.7 \\
Think w/o CoT & 27.7 & 44.2 & 71.4 \\
\midrule
RL-Zero-Math & 14.2 & 34.9 & 59.8 \\
RL-Zero-Code & 10.0 & 27.0 & 52.2 \\
RL-Zero-IF & 23.0 & 32.9 & 58.8 \\
RL-Zero-General & 18.8 & 37.9 & 68.2 \\
RL-Zero-Math$^{3.1}$ & 9.8 & 25.8 & 50.5 \\
RL-Zero-Code$^{3.1}$ & 10.8 & 28.4 & 55.0 \\
\bottomrule
\end{tabular}
\caption{CruxEval output prediction quality (\%). Accuracy@1, majority vote@16, and pass@16.}
\label{tab:app_quality_cruxeval}

\end{table}

%% file: tables/app_quality_summarization.tex
\begin{table}[htbp]
\centering
\footnotesize
\setlength{\tabcolsep}{5pt}
\resizebox{0.6\textwidth}{!}{%
\begin{tabular}{lccc}
\toprule
 & TL;DR & CNN/DM & XSum \\
\midrule
Base & 26.0 & 47.7 & 26.7 \\
\midrule
Instruct-SFT & 72.2 & 20.0 & 52.0 \\
Instruct-DPO & 70.4 & 95.4 & 94.4 \\
Instruct (final) & \textbf{77.8} & 95.4 & 95.4 \\
\midrule
Think-SFT & 32.0 & 97.2 & 95.8 \\
Think-DPO & 28.4 & 91.6 & 90.6 \\
Think (final) & 38.2 & \textbf{97.8} & \textbf{96.0} \\
\midrule
Think-SFT w/o CoT & 20.0 & 55.6 & 78.4 \\
Think-DPO w/o CoT & 13.2 & 44.7 & 67.8 \\
Think w/o CoT & 12.5 & 49.6 & 73.9 \\
\midrule
RL-Zero-Math & 35.4 & 49.3 & 37.4 \\
RL-Zero-Code & 37.2 & 49.0 & 41.4 \\
RL-Zero-IF & 36.8 & 41.6 & 39.4 \\
RL-Zero-General & 44.0 & 60.6 & 43.9 \\
RL-Zero-Math$^{3.1}$ & 34.4 & 50.0 & 39.8 \\
RL-Zero-Code$^{3.1}$ & 33.0 & 56.2 & 40.6 \\
\bottomrule
\end{tabular}}
\caption{Summarization quality: pairwise win rate (\%) against reference summaries, judged by gpt-4.1-mini.}
\label{tab:app_quality_summarization}

\end{table}

%% file: tables/app_quality_openended.tex
\begin{table}[htbp]
\centering
\footnotesize
\setlength{\tabcolsep}{5pt}
\resizebox{0.7\textwidth}{!}{%
\begin{tabular}{lccc}
\toprule
 & Alpaca & PRISM & WritingPrompts \\
\midrule
Base & --- & --- & --- \\
\midrule
Instruct-SFT & 48.7 & 84.0 & 93.1 \\
Instruct-DPO & 73.9 & 93.1 & 96.9 \\
Instruct (final) & 66.3 & 91.0 & 97.3 \\
\midrule
Think-SFT & 84.3 & 92.5 & 96.8 \\
Think-DPO & \textbf{95.2} & \textbf{95.7} & \textbf{98.0} \\
Think (final) & 83.9 & 93.7 & 97.3 \\
\midrule
Think-SFT w/o CoT & 83.4 & 88.6 & 92.5 \\
Think-DPO w/o CoT & 88.0 & 92.0 & 95.4 \\
Think w/o CoT & 84.7 & 88.6 & 93.3 \\
\midrule
RL-Zero-Math & 53.1 & 51.7 & 49.9 \\
RL-Zero-Code & 55.4 & 61.9 & 54.6 \\
RL-Zero-IF & 32.8 & 53.3 & 52.4 \\
RL-Zero-General & 76.7 & 63.0 & 53.6 \\
RL-Zero-Math$^{3.1}$ & 57.5 & 55.7 & 49.2 \\
RL-Zero-Code$^{3.1}$ & 52.5 & 52.1 & 49.4 \\
\bottomrule
\end{tabular}}
\caption{Open-ended quality: pairwise win rate (\%) against Base model. Alpaca and PRISM use the MT-Bench pair-v2 prompt \citep{zheng2023judging}; WritingPrompts uses the Arena-Hard creative writing prompt \citep{li2024arenahard} with position-swap debiasing. Judge: gpt-4.1-mini.}
\label{tab:app_quality_openended}
\end{table}

%% file: tables/app_quality_wildbench.tex
\begin{table}[htbp]
\centering
\footnotesize
\setlength{\tabcolsep}{5pt}
\resizebox{0.6\textwidth}{!}{%
\begin{tabular}{lcccc}
\toprule
 & Raw & $\sigma$ & Median & WB-Score \\
\midrule
Base & 4.0 & 2.4 & 4 & -2.0 \\
\midrule
Instruct-SFT & 7.2 & 1.9 & 8 & 4.5 \\
Instruct-DPO & 7.6 & 1.7 & 8 & 5.2 \\
Instruct (final) & \textbf{8.0} & 1.5 & \textbf{9} & \textbf{6.1} \\
\midrule
Think-SFT & 7.2 & 2.1 & 8 & 4.3 \\
Think-DPO & 7.5 & 2.0 & 8 & 5.1 \\
Think (final) & 7.3 & 2.0 & 8 & 4.6 \\
\midrule
Think-SFT w/o CoT & 5.4 & 2.6 & 5 & 0.8 \\
Think-DPO w/o CoT & 5.7 & 2.3 & 6 & 1.4 \\
Think w/o CoT & 5.7 & 2.5 & 6 & 1.4 \\
\midrule
RL-Zero-Math & 4.1 & 2.5 & 4 & -1.7 \\
RL-Zero-Code & 4.2 & 2.6 & 4 & -1.6 \\ 
RL-Zero-IF & 4.0 & 2.5 & 4 & -2.0 \\
RL-Zero-General & 4.9 & 2.7 & 5 & -0.2 \\
RL-Zero-Math$^{3.1}$ & 4.0 & 2.5 & 4 & -2.0 \\
RL-Zero-Code$^{3.1}$ & 4.2 & 2.6 & 4 & -1.6 \\
\bottomrule
\end{tabular}}
\caption{WildBench quality: checklist-guided WB-Score \citep{lin2025wildbench}, judged by gpt-4.1-mini. Raw score (1--10) and normalized WB-Score $= (\text{raw} - 5) \times 2$.}
\label{tab:app_quality_wildbench}
\end{table}

%% file: tables/app_qfd.tex
\begin{table*}[htbp]
\centering
\footnotesize
\setlength{\tabcolsep}{3pt}
\resizebox{\textwidth}{!}{%
\begin{tabular}{lcccccccccccc}
\toprule
 & \multicolumn{4}{c}{GSM8K} & \multicolumn{4}{c}{MATH-Algebra} & \multicolumn{4}{c}{MATH-Geometry} \\
\cmidrule(lr){2-5}\cmidrule(lr){6-9}\cmidrule(lr){10-13}
 & acc & $D_a$ & $D_c$ & $V_c$ & acc & $D_a$ & $D_c$ & $V_c$ & acc & $D_a$ & $D_c$ & $V_c$ \\
\midrule
Base & 52 & 0.172 & 0.135 & 1.7 & 48 & 0.146 & 0.119 & 1.6 & 23 & 0.198 & 0.145 & 1.6 \\
\midrule
Instruct-SFT & 73 & 0.105 & 0.098 & 1.5 & 56 & 0.132 & 0.110 & 1.6 & 26 & 0.179 & 0.140 & 1.6 \\
Instruct-DPO & 77 & 0.141 & 0.137 & 1.8 & 51 & 0.071 & 0.067 & 1.4 & 23 & 0.096 & 0.082 & 1.4 \\
Instruct (final) & 80 & 0.078 & 0.074 & 1.4 & 71 & 0.057 & 0.057 & 1.4 & 43 & 0.101 & 0.087 & 1.5 \\
\midrule
Think-SFT & 92 & 0.061 & 0.060 & 1.4 & 76 & 0.054 & 0.051 & 1.3 & 50 & 0.107 & 0.080 & 1.5 \\
Think-DPO & 85 & 0.052 & 0.049 & 1.3 & 75 & 0.061 & 0.053 & 1.3 & 50 & 0.114 & 0.082 & 1.5 \\
Think (final) & \textbf{93} & 0.051 & 0.050 & 1.3 & \textbf{77} & 0.062 & 0.059 & 1.4 & \textbf{51} & 0.122 & 0.091 & 1.6 \\
\midrule
Think-SFT w/o CoT & 77 & 0.057 & 0.055 & 1.3 & 56 & 0.066 & 0.058 & 1.3 & 28 & 0.098 & 0.072 & 1.4 \\
Think-DPO w/o CoT & 70 & 0.045 & 0.042 & 1.2 & 47 & 0.058 & 0.050 & 1.3 & 20 & 0.077 & 0.061 & 1.3 \\
Think w/o CoT & 75 & 0.052 & 0.048 & 1.3 & 49 & 0.064 & 0.055 & 1.3 & 20 & 0.089 & 0.064 & 1.3 \\
\midrule
RL-Zero-Math & 61 & 0.154 & 0.124 & 1.7 & 49 & 0.144 & 0.119 & 1.6 & 23 & 0.181 & 0.135 & 1.6 \\
RL-Zero-Code & 58 & 0.156 & 0.127 & 1.7 & 51 & 0.144 & 0.114 & 1.6 & 23 & 0.183 & 0.135 & 1.6 \\
RL-Zero-IF & 50 & 0.177 & 0.137 & 1.7 & 48 & 0.143 & 0.111 & 1.6 & 21 & 0.199 & 0.132 & 1.6 \\
RL-Zero-General & 61 & 0.133 & 0.110 & 1.6 & 54 & 0.124 & 0.104 & 1.6 & 24 & 0.166 & 0.127 & 1.5 \\
RL-Zero-Math$^{3.1}$ & 55 & 0.183 & 0.136 & 1.7 & 54 & 0.140 & 0.120 & 1.6 & 21 & 0.183 & 0.133 & 1.6 \\
RL-Zero-Code$^{3.1}$ & 60 & 0.173 & 0.130 & 1.7 & 52 & 0.139 & 0.115 & 1.6 & 22 & 0.178 & 0.133 & 1.6 \\
\bottomrule
\end{tabular}}
\vspace{4pt}
\resizebox{\textwidth}{!}{%
\begin{tabular}{lcccccccccccccccc}
\toprule
 & \multicolumn{4}{c}{IFEval} & \multicolumn{4}{c}{HumanEval} & \multicolumn{4}{c}{MBPP} & \multicolumn{4}{c}{CRUXEval} \\
\cmidrule(lr){2-5}\cmidrule(lr){6-9}\cmidrule(lr){10-13}\cmidrule(lr){14-17}
 & acc & $D_a$ & $D_c$ & $V_c$ & acc & $D_a$ & $D_c$ & $V_c$ & acc & $D_a$ & $D_c$ & $V_c$ & acc & $D_a$ & $D_c$ & $V_c$ \\
\midrule
Base & 45 & 0.349 & 0.333 & 3.2 & 18 & 0.411 & 0.123 & 1.5 & 19 & 0.291 & 0.196 & 1.9 & 20 & 0.239 & 0.240 & 1.9 \\
\midrule
Instruct-SFT & 79 & 0.172 & 0.171 & 2.2 & 63 & 0.112 & 0.109 & 1.6 & 32 & 0.111 & 0.098 & 1.5 & 32 & 0.218 & 0.177 & 1.7 \\
Instruct-DPO & 79 & 0.154 & 0.155 & 2.1 & 73 & 0.095 & 0.095 & 1.6 & 33 & 0.073 & 0.059 & 1.3 & \textbf{38} & 0.068 & 0.168 & 1.6 \\
Instruct (final) & \textbf{82} & 0.154 & 0.155 & 2.1 & 81 & 0.093 & 0.091 & 1.6 & 38 & 0.069 & 0.058 & 1.3 & 23 & 0.062 & 0.139 & 1.4 \\
\midrule
Think-SFT & 78 & 0.191 & 0.180 & 2.3 & 87 & 0.109 & 0.101 & 1.6 & 41 & 0.081 & 0.058 & 1.3 & 18 & 0.095 & 0.076 & 1.3 \\
Think-DPO & 75 & 0.165 & 0.159 & 2.1 & 87 & 0.081 & 0.072 & 1.4 & 36 & 0.084 & 0.067 & 1.4 & 17 & 0.076 & 0.056 & 1.2 \\
Think (final) & 79 & 0.196 & 0.187 & 2.3 & \textbf{88} & 0.117 & 0.110 & 1.6 & \textbf{44} & 0.089 & 0.064 & 1.4 & 12 & 0.090 & 0.074 & 1.3 \\
\midrule
Think-SFT w/o CoT & 70 & 0.196 & 0.185 & 2.2 & 49 & 0.055 & 0.046 & 1.3 & 24 & 0.084 & 0.072 & 1.4 & 20 & 0.084 & 0.087 & 1.4 \\
Think-DPO w/o CoT & 67 & 0.157 & 0.152 & 2.0 & 56 & 0.062 & 0.051 & 1.3 & 26 & 0.083 & 0.065 & 1.3 & 21 & 0.064 & 0.098 & 1.4 \\
Think w/o CoT & 71 & 0.221 & 0.182 & 2.1 & 56 & 0.060 & 0.053 & 1.3 & 24 & 0.083 & 0.070 & 1.4 & 20 & 0.071 & 0.081 & 1.3 \\
\midrule
RL-Zero-Math & 48 & 0.318 & 0.295 & 2.9 & 3 & 0.421 & 0.089 & 1.4 & 24 & 0.274 & 0.157 & 1.8 & 18 & 0.222 & 0.245 & 1.9 \\
RL-Zero-Code & 47 & 0.287 & 0.278 & 2.7 & 3 & 0.464 & 0.180 & 1.5 & 25 & 0.238 & 0.147 & 1.7 & 16 & 0.149 & 0.201 & 1.7 \\
RL-Zero-IF & 60 & 0.397 & 0.371 & 3.9 & 0 & 0.336 & -- & -- & 24 & 0.297 & 0.149 & 1.7 & 24 & 0.491 & 0.319 & 2.1 \\
RL-Zero-General & 47 & 0.284 & 0.271 & 2.7 & 32 & 0.468 & 0.113 & 1.5 & 25 & 0.272 & 0.151 & 1.7 & 23 & 0.198 & 0.190 & 1.8 \\
RL-Zero-Math$^{3.1}$ & 49 & 0.324 & 0.300 & 2.9 & 7 & 0.460 & 0.116 & 1.4 & 24 & 0.292 & 0.157 & 1.8 & 19 & 0.207 & 0.236 & 1.8 \\
RL-Zero-Code$^{3.1}$ & 46 & 0.325 & 0.293 & 2.8 & 66 & 0.439 & 0.071 & 1.4 & 26 & 0.261 & 0.153 & 1.8 & 17 & 0.209 & 0.247 & 1.8 \\
\bottomrule
\end{tabular}
}
\caption{Quality-filtered diversity. acc: accuracy (\%). $D_a$: SBERT on all outputs. $D_c$: SBERT on correct only ($K_c \geq 2$). $V_c$: Vendi Score on correct only (effective number of distinct answers).}
\label{tab:app_qfd}
\end{table*}

%% file: tables/app_code_diversity.tex
\begin{table*}[htbp]
\centering
\footnotesize
\setlength{\tabcolsep}{3pt}
\begin{tabular}{lrrr rrr}
\toprule
 & \multicolumn{3}{c}{HumanEval} & \multicolumn{3}{c}{MBPP} \\
\cmidrule(lr){2-4}\cmidrule(lr){5-7}
 & acc & $D_c^{\text{code}}$ & $D_c^{\text{AST}}$ & acc & $D_c^{\text{code}}$ & $D_c^{\text{AST}}$ \\
\midrule
Base & 18 & 0.168 & 0.590 & 19 & 0.310 & 0.927 \\
\midrule
Instruct-SFT & 63 & 0.167 & 0.591 & 32 & 0.179 & 0.683 \\
Instruct-DPO & 74 & 0.113 & 0.674 & 33 & 0.118 & 0.777 \\
Instruct (final) & 81 & 0.142 & 0.593 & 38 & 0.118 & 0.693 \\
\midrule
Think-SFT & \textbf{91} & 0.116 & 0.527 & 40 & 0.124 & 0.533 \\
Think-DPO & \textbf{91} & 0.130 & 0.540 & 39 & 0.162 & 0.611 \\
Think (final) & \textbf{91} & 0.126 & 0.531 & \textbf{44} & 0.130 & 0.590 \\
\midrule
Think-SFT w/o CoT & 51 & 0.105 & 0.510 & 26 & 0.170 & 0.662 \\
Think-DPO w/o CoT & 59 & 0.123 & 0.485 & 28 & 0.160 & 0.707 \\
Think w/o CoT & 57 & 0.112 & 0.499 & 26 & 0.164 & 0.676 \\
\midrule
RL-Zero-Math & 3 & 0.058 & 0.057 & 25 & 0.253 & 0.905 \\
RL-Zero-Code & 3 & 0.254 & --- & 25 & 0.249 & 0.889 \\
RL-Zero-IF & 0 & --- & --- & 25 & 0.245 & 0.887 \\
RL-Zero-General & 33 & 0.174 & 0.618 & 25 & 0.255 & 0.900 \\
RL-Zero-Math$^{3.1}$ & 7 & 0.101 & 0.261 & 24 & 0.263 & 0.889 \\
RL-Zero-Code$^{3.1}$ & 67 & 0.124 & 0.576 & 25 & 0.249 & 0.895 \\
\bottomrule
\end{tabular}
\caption{Code-specific diversity on correct outputs for code-generation tasks. acc: accuracy (\%, mean $K_c$/16). $D_c^{\text{code}}$: UniXcoder SBERT (correct only). $D_c^{\text{AST}}$: AST subtree Jaccard (correct only).}
\label{tab:code_diversity}
\end{table*}

%% file: tables/app_temperature.tex
\begin{table*}[t]
\centering
\footnotesize
\setlength{\tabcolsep}{2pt}
\begin{tabular}{lrrrrrrrrrrrr}
\toprule
 & \multicolumn{3}{c}{\textbf{EAD}} & \multicolumn{3}{c}{\textbf{SBERT}} & \multicolumn{3}{c}{\textbf{NLI}} & \multicolumn{3}{c}{\textbf{Vendi}} \\
\cmidrule(lr){2-4} \cmidrule(lr){5-7} \cmidrule(lr){8-10} \cmidrule(lr){11-13}
Task & $T{=}1.0$ & $T{=}0.6$ & $\Delta$\% & $T{=}1.0$ & $T{=}0.6$ & $\Delta$\% & $T{=}1.0$ & $T{=}0.6$ & $\Delta$\% & $T{=}1.0$ & $T{=}0.6$ & $\Delta$\% \\
\midrule
TL;DR & 0.478 & 0.365 & -23.6 & 0.385 & 0.353 & -8.1 & 0.987 & 0.949 & -3.8 & 4.556 & 4.157 & -8.8 \\
CNN/DM & 0.439 & 0.370 & -15.8 & 0.254 & 0.279 & +9.8 & 0.973 & 1.036 & +6.5 & 2.928 & 3.162 & +8.0 \\
XSum & 0.743 & 0.674 & -9.3 & 0.551 & 0.451 & -18.2 & 1.122 & 1.087 & -3.1 & 6.628 & 5.192 & -21.7 \\
\midrule
HumanEval & 0.623 & 0.570 & -8.5 & 0.438 & 0.411 & -6.3 & 1.037 & 0.894 & -13.7 & 2.578 & 2.454 & -4.8 \\
MBPP & 0.631 & 0.592 & -6.2 & 0.431 & 0.291 & -32.5 & 1.236 & 1.179 & -4.7 & 3.479 & 2.753 & -20.9 \\
CRUXEval & 0.429 & 0.310 & -27.6 & 0.293 & 0.239 & -18.2 & 0.992 & 0.997 & +0.6 & 2.841 & 2.458 & -13.5 \\
\midrule
GSM8K & 0.433 & 0.450 & +3.9 & 0.199 & 0.172 & -13.6 & 1.095 & 1.077 & -1.7 & 2.251 & 2.094 & -7.0 \\
MATH-Algebra & 0.472 & 0.453 & -3.9 & 0.156 & 0.146 & -6.2 & 1.012 & 1.000 & -1.2 & 2.050 & 1.983 & -3.2 \\
MATH-Geometry & 0.476 & 0.402 & -15.5 & 0.210 & 0.198 & -5.9 & 1.114 & 1.135 & +1.8 & 2.525 & 2.438 & -3.5 \\
TruthfulQA & 0.616 & 0.461 & -25.2 & 0.452 & 0.353 & -21.8 & 1.097 & 0.972 & -11.4 & 5.173 & 3.805 & -26.4 \\
\midrule
Alpaca & 0.539 & 0.509 & -5.6 & 0.396 & 0.319 & -19.5 & 0.791 & 0.676 & -14.5 & 2.620 & 2.217 & -15.4 \\
IFEval & 0.554 & 0.443 & -20.1 & 0.371 & 0.349 & -6.0 & 1.063 & 1.055 & -0.7 & 4.134 & 3.812 & -7.8 \\
\midrule
WritingPrompts & 0.357 & 0.234 & -34.5 & 0.588 & 0.540 & -8.1 & 1.166 & 1.165 & -0.1 & 7.914 & 6.935 & -12.4 \\
PRISM & 0.376 & 0.237 & -37.0 & 0.452 & 0.408 & -9.7 & 1.101 & 1.086 & -1.4 & 5.313 & 4.639 & -12.7 \\
WildBench & 0.475 & 0.300 & -36.9 & 0.350 & 0.335 & -4.3 & 1.087 & 1.064 & -2.1 & 3.702 & 3.514 & -5.1 \\
\midrule
\textbf{Mean} & 0.509 & 0.425 & -17.7 & 0.368 & 0.323 & -11.2 & 1.058 & 1.025 & -3.3 & 3.913 & 3.441 & -10.3 \\
\bottomrule
\end{tabular}
\caption{Base model diversity at recommended ($T{=}1.0$) vs.\ matched ($T{=}0.6$) temperature. $\Delta$\% reports the relative change.}
\label{tab:app_temperature}
\end{table*}

%% file: tables/stage_attribution_full.tex
\begin{table*}[t]
\centering\small
\setlength{\tabcolsep}{3pt}
\begin{tabular}{l rrr r rrr r r}
\toprule
& \multicolumn{4}{c}{Think} & \multicolumn{4}{c}{Instruct} & RL-Zero \\
\cmidrule(lr){2-5} \cmidrule(lr){6-9} \cmidrule(lr){10-10}
Task & SFT & DPO & RL & Retain & SFT & DPO & RL & Retain & Retain \\
\midrule
TL;DR & $-$53 & $-$2 & $+$1 & 46 & $-$24 & $-$19 & $+$1 & 59 & 93 \\
CNN/DM & $-$70 & $-$8 & $+$11 & 33 & $-$20 & $-$53 & $-$1 & 26 & 71 \\
XSum & $-$80 & $-$6 & $+$6 & 20 & $-$37 & $-$44 & $-$0 & 18 & 94 \\
HumanEval & $-$73 & $-$7 & $+$9 & 28 & $-$73 & $-$4 & $-$0 & 23 & 105 \\
MBPP & $-$72 & $+$1 & $+$1 & 31 & $-$62 & $-$13 & $-$2 & 24 & 94 \\
CRUXEval & $-$60 & $-$8 & $+$6 & 38 & $-$9 & $-$63 & $-$3 & 26 & 103 \\
GSM8K & $-$64 & $-$6 & $-$0 & 30 & $-$39 & $+$21 & $-$37 & 45 & 95 \\
MATH-Alg & $-$63 & $+$5 & $+$1 & 43 & $-$10 & $-$42 & $-$9 & 39 & 95 \\
MATH-Geo & $-$46 & $+$3 & $+$4 & 62 & $-$9 & $-$42 & $+$2 & 51 & 92 \\
IFEval & $-$45 & $-$7 & $+$9 & 56 & $-$51 & $-$5 & $+$0 & 44 & 92 \\
Alpaca & $-$56 & $-$7 & $+$9 & 46 & $-$47 & $-$15 & $-$2 & 35 & 76 \\
WritingPrompts & $-$56 & $-$6 & $-$1 & 37 & $-$49 & $-$9 & $-$4 & 37 & 100 \\
TruthfulQA & $-$66 & $-$13 & $+$0 & 21 & $-$8 & $-$48 & $-$12 & 33 & 99 \\
PRISM & $-$75 & $-$3 & $+$1 & 22 & $-$65 & $-$11 & $-$1 & 22 & 95 \\
WildBench & $-$52 & $-$2 & $+$6 & 52 & $-$61 & $-$2 & $-$1 & 35 & 89 \\
\midrule
\textbf{Average} & \textbf{$-$62} & \textbf{$-$4} & \textbf{$+$4} & \textbf{38} & \textbf{$-$38} & \textbf{$-$23} & \textbf{$-$5} & \textbf{34} & \textbf{93} \\
\bottomrule
\end{tabular}
\caption{Per-task stage attribution: percentage of Base SBERT diversity lost ($-$) or recovered ($+$) at each post-training stage. \emph{Retain} is the fraction of Base diversity preserved at the final checkpoint.}
\label{tab:stage_attribution_full}
\end{table*}

%% file: tables/decontamination.tex
\begin{table}[t]
\centering
\footnotesize
\setlength{\tabcolsep}{2pt}
\resizebox{\textwidth}{!}{%
\begin{tabular}{lccccccccccccccc}
\toprule
& \rotatebox{70}{CNN/DM}
& \rotatebox{70}{XSum}
& \rotatebox{70}{TL;DR}
& \rotatebox{70}{GSM8K}
& \rotatebox{70}{MATH-Alg}
& \rotatebox{70}{MATH-Geo}
& \rotatebox{70}{HumanEval}
& \rotatebox{70}{MBPP}
& \rotatebox{70}{CRUXEval}
& \rotatebox{70}{Alpaca}
& \rotatebox{70}{IFEval}
& \rotatebox{70}{WrtPrompts}
& \rotatebox{70}{TruthfulQA}
& \rotatebox{70}{PRISM}
& \rotatebox{70}{WildBench} \\
\midrule
Think-SFT & 0.3 & 0.2 & 0.1 & 0.2 & \textbf{10.2} & \textbf{15.4} & \textbf{21.5} & 1.1 & \textbf{9.5} & 0.4 & \textbf{7.6} & 0.0 & 0.0 & 0.0 & \textbf{20.5} \\
Think-DPO & 0.2 & 0.3 & 0.0 & 0.0 & 1.2 & 1.2 & \textbf{14.7} & 0.0 & \textbf{5.5} & 0.7 & \textbf{6.8} & 0.0 & 0.0 & 0.0 & \textbf{6.5} \\
Inst.-SFT & 0.1 & 0.5 & 0.1 & 0.0 & 2.7 & 3.0 & \textbf{30.1} & 1.3 & \textbf{11.2} & 0.4 & \textbf{7.3} & 0.0 & 0.0 & 0.0 & \textbf{26.4} \\
Inst.-DPO & 0.1 & 0.1 & 0.0 & 0.0 & 1.8 & 1.8 & \textbf{20.9} & 0.3 & \textbf{6.8} & 1.6 & \textbf{7.2} & 0.0 & 0.0 & 0.0 & \textbf{6.9} \\
\bottomrule
\end{tabular}
}
\caption{$C_{13}$ 13-gram overlap (\%) between Dolci training sets and evaluation benchmarks. Values ${\ge}\,5\%$ are \textbf{bolded}.}
\label{tab:decontamination}
\end{table}

%% file: references.bib
@inproceedings{
rafailov2023direct,
title={Direct Preference Optimization: Your Language Model is Secretly a Reward Model},
author={Rafael Rafailov and Archit Sharma and Eric Mitchell and Christopher D Manning and Stefano Ermon and Chelsea Finn},
booktitle={Thirty-seventh Conference on Neural Information Processing Systems},
year={2023},
url={https://openreview.net/forum?id=HPuSIXJaa9}
}

@inproceedings{
wang2024beyond,
title={Beyond Reverse {KL}: Generalizing Direct Preference Optimization with Diverse Divergence Constraints},
author={Chaoqi Wang and Yibo Jiang and Chenghao Yang and Han Liu and Yuxin Chen},
booktitle={The Twelfth International Conference on Learning Representations},
year={2024},
url={https://openreview.net/forum?id=2cRzmWXK9N}
}

@inproceedings{
kirk2024understanding,
title={Understanding the Effects of {RLHF} on {LLM} Generalisation and Diversity},
author={Robert Kirk and Ishita Mediratta and Christoforos Nalmpantis and Jelena Luketina and Eric Hambro and Edward Grefenstette and Roberta Raileanu},
booktitle={The Twelfth International Conference on Learning Representations},
year={2024},
url={https://openreview.net/forum?id=PXD3FAVHJT}
}

@article{karouzos2026empirical,
  title={An Empirical Study on Preference Tuning Generalization and Diversity Under Domain Shift},
  author={Karouzos, Constantinos and Tan, Xingwei and Aletras, Nikolaos},
  journal={arXiv preprint arXiv:2601.05882},
  year={2026}
}

@inproceedings{yun-etal-2025-price,
    title = "The Price of Format: Diversity Collapse in {LLM}s",
    author = "Yun, Longfei  and
      An, Chenyang  and
      Wang, Zilong  and
      Peng, Letian  and
      Shang, Jingbo",
    editor = "Christodoulopoulos, Christos  and
      Chakraborty, Tanmoy  and
      Rose, Carolyn  and
      Peng, Violet",
    booktitle = "Findings of the Association for Computational Linguistics: EMNLP 2025",
    month = nov,
    year = "2025",
    address = "Suzhou, China",
    publisher = "Association for Computational Linguistics",
    url = "https://aclanthology.org/2025.findings-emnlp.836/",
    doi = "10.18653/v1/2025.findings-emnlp.836",
    pages = "15454--15468",
    ISBN = "979-8-89176-335-7"
}

@inproceedings{
razin2025unintentional,
title={Unintentional Unalignment: Likelihood Displacement in Direct Preference Optimization},
author={Noam Razin and Sadhika Malladi and Adithya Bhaskar and Danqi Chen and Sanjeev Arora and Boris Hanin},
booktitle={The Thirteenth International Conference on Learning Representations},
year={2025},
url={https://openreview.net/forum?id=uaMSBJDnRv}
}

@misc{ma2025gradient,
      title={Gradient Imbalance in Direct Preference Optimization}, 
      author={Qinwei Ma and Jingzhe Shi and Can Jin and Jenq-Neng Hwang and Serge Belongie and Lei Li},
      year={2025},
      eprint={2502.20847},
      archivePrefix={arXiv},
      primaryClass={cs.LG},
      url={https://arxiv.org/abs/2502.20847}, 
}

@inproceedings{
gxchen2025kl,
title={{KL}-Regularized Reinforcement Learning is Designed to Mode Collapse},
author={Anthony GX-Chen and Jatin Prakash and Jeff Guo and Rob Fergus and Rajesh Ranganath},
booktitle={The Fourteenth International Conference on Learning Representations},
year={2026},
url={https://openreview.net/forum?id=flBRtdIihA}
}

@inproceedings{
slocum2025diverse,
title={Diverse Preference Learning for Capabilities and Alignment},
author={Stewart Slocum and Asher Parker-Sartori and Dylan Hadfield-Menell},
booktitle={The Thirteenth International Conference on Learning Representations},
year={2025},
url={https://openreview.net/forum?id=pOq9vDIYev}
}

@misc{pan2026qempo,
      title={Quality-constrained Entropy Maximization Policy Optimization for LLM Diversity}, 
      author={Haihui Pan and Yuzhong Hong and Shaoke Lv and Junwei Bao and Hongfei Jiang and Yang Song},
      year={2026},
      eprint={2602.15894},
      archivePrefix={arXiv},
      primaryClass={cs.CL},
      url={https://arxiv.org/abs/2602.15894}, 
}

@misc{lanchantin2025divpo,
      title={Diverse Preference Optimization}, 
      author={Jack Lanchantin and Angelica Chen and Shehzaad Dhuliawala and Ping Yu and Jason Weston and Sainbayar Sukhbaatar and Ilia Kulikov},
      year={2025},
      eprint={2501.18101},
      archivePrefix={arXiv},
      primaryClass={cs.CL},
      url={https://arxiv.org/abs/2501.18101}, 
}

@misc{olmo2025olmo3,
      title={Olmo 3}, 
      author={Team Olmo and : and Allyson Ettinger and Amanda Bertsch and Bailey Kuehl and David Graham and David Heineman and Dirk Groeneveld and Faeze Brahman and Finbarr Timbers and Hamish Ivison and Jacob Morrison and Jake Poznanski and Kyle Lo and Luca Soldaini and Matt Jordan and Mayee Chen and Michael Noukhovitch and Nathan Lambert and Pete Walsh and Pradeep Dasigi and Robert Berry and Saumya Malik and Saurabh Shah and Scott Geng and Shane Arora and Shashank Gupta and Taira Anderson and Teng Xiao and Tyler Murray and Tyler Romero and Victoria Graf and Akari Asai and Akshita Bhagia and Alexander Wettig and Alisa Liu and Aman Rangapur and Chloe Anastasiades and Costa Huang and Dustin Schwenk and Harsh Trivedi and Ian Magnusson and Jaron Lochner and Jiacheng Liu and Lester James V. Miranda and Maarten Sap and Malia Morgan and Michael Schmitz and Michal Guerquin and Michael Wilson and Regan Huff and Ronan Le Bras and Rui Xin and Rulin Shao and Sam Skjonsberg and Shannon Zejiang Shen and Shuyue Stella Li and Tucker Wilde and Valentina Pyatkin and Will Merrill and Yapei Chang and Yuling Gu and Zhiyuan Zeng and Ashish Sabharwal and Luke Zettlemoyer and Pang Wei Koh and Ali Farhadi and Noah A. Smith and Hannaneh Hajishirzi},
      year={2025},
      eprint={2512.13961},
      archivePrefix={arXiv},
      primaryClass={cs.CL},
      url={https://arxiv.org/abs/2512.13961}, 
}

@article{deepseek2025r1,
   title={DeepSeek-R1 incentivizes reasoning in LLMs through reinforcement learning},
   volume={645},
   ISSN={1476-4687},
   url={http://dx.doi.org/10.1038/s41586-025-09422-z},
   DOI={10.1038/s41586-025-09422-z},
   number={8081},
   journal={Nature},
   publisher={Springer Science and Business Media LLC},
   author={Guo, Daya and Yang, Dejian and Zhang, Haowei and Song, Junxiao and Wang, Peiyi and Zhu, Qihao and Xu, Runxin and Zhang, Ruoyu and Ma, Shirong and Bi, Xiao and Zhang, Xiaokang and Yu, Xingkai and Wu, Yu and Wu, Z. F. and Gou, Zhibin and Shao, Zhihong and Li, Zhuoshu and Gao, Ziyi and Liu, Aixin and Xue, Bing and Wang, Bingxuan and Wu, Bochao and Feng, Bei and Lu, Chengda and Zhao, Chenggang and Deng, Chengqi and Ruan, Chong and Dai, Damai and Chen, Deli and Ji, Dongjie and Li, Erhang and Lin, Fangyun and Dai, Fucong and Luo, Fuli and Hao, Guangbo and Chen, Guanting and Li, Guowei and Zhang, H. and Xu, Hanwei and Ding, Honghui and Gao, Huazuo and Qu, Hui and Li, Hui and Guo, Jianzhong and Li, Jiashi and Chen, Jingchang and Yuan, Jingyang and Tu, Jinhao and Qiu, Junjie and Li, Junlong and Cai, J. L. and Ni, Jiaqi and Liang, Jian and Chen, Jin and Dong, Kai and Hu, Kai and You, Kaichao and Gao, Kaige and Guan, Kang and Huang, Kexin and Yu, Kuai and Wang, Lean and Zhang, Lecong and Zhao, Liang and Wang, Litong and Zhang, Liyue and Xu, Lei and Xia, Leyi and Zhang, Mingchuan and Zhang, Minghua and Tang, Minghui and Zhou, Mingxu and Li, Meng and Wang, Miaojun and Li, Mingming and Tian, Ning and Huang, Panpan and Zhang, Peng and Wang, Qiancheng and Chen, Qinyu and Du, Qiushi and Ge, Ruiqi and Zhang, Ruisong and Pan, Ruizhe and Wang, Runji and Chen, R. J. and Jin, R. L. and Chen, Ruyi and Lu, Shanghao and Zhou, Shangyan and Chen, Shanhuang and Ye, Shengfeng and Wang, Shiyu and Yu, Shuiping and Zhou, Shunfeng and Pan, Shuting and Li, S. S. and Zhou, Shuang and Wu, Shaoqing and Yun, Tao and Pei, Tian and Sun, Tianyu and Wang, T. and Zeng, Wangding and Liu, Wen and Liang, Wenfeng and Gao, Wenjun and Yu, Wenqin and Zhang, Wentao and Xiao, W. L. and An, Wei and Liu, Xiaodong and Wang, Xiaohan and Chen, Xiaokang and Nie, Xiaotao and Cheng, Xin and Liu, Xin and Xie, Xin and Liu, Xingchao and Yang, Xinyu and Li, Xinyuan and Su, Xuecheng and Lin, Xuheng and Li, X. Q. and Jin, Xiangyue and Shen, Xiaojin and Chen, Xiaosha and Sun, Xiaowen and Wang, Xiaoxiang and Song, Xinnan and Zhou, Xinyi and Wang, Xianzu and Shan, Xinxia and Li, Y. K. and Wang, Y. Q. and Wei, Y. X. and Zhang, Yang and Xu, Yanhong and Li, Yao and Zhao, Yao and Sun, Yaofeng and Wang, Yaohui and Yu, Yi and Zhang, Yichao and Shi, Yifan and Xiong, Yiliang and He, Ying and Piao, Yishi and Wang, Yisong and Tan, Yixuan and Ma, Yiyang and Liu, Yiyuan and Guo, Yongqiang and Ou, Yuan and Wang, Yuduan and Gong, Yue and Zou, Yuheng and He, Yujia and Xiong, Yunfan and Luo, Yuxiang and You, Yuxiang and Liu, Yuxuan and Zhou, Yuyang and Zhu, Y. X. and Huang, Yanping and Li, Yaohui and Zheng, Yi and Zhu, Yuchen and Ma, Yunxian and Tang, Ying and Zha, Yukun and Yan, Yuting and Ren, Z. Z. and Ren, Zehui and Sha, Zhangli and Fu, Zhe and Xu, Zhean and Xie, Zhenda and Zhang, Zhengyan and Hao, Zhewen and Ma, Zhicheng and Yan, Zhigang and Wu, Zhiyu and Gu, Zihui and Zhu, Zijia and Liu, Zijun and Li, Zilin and Xie, Ziwei and Song, Ziyang and Pan, Zizheng and Huang, Zhen and Xu, Zhipeng and Zhang, Zhongyu and Zhang, Zhen},
   year={2025},
   month=sep, pages={633–638} }

@inproceedings{liu-etal-2022-rethinking,
    title = "Rethinking and Refining the Distinct Metric",
    author = "Liu, Siyang  and
      Sabour, Sahand  and
      Zheng, Yinhe  and
      Ke, Pei  and
      Zhu, Xiaoyan  and
      Huang, Minlie",
    editor = "Muresan, Smaranda  and
      Nakov, Preslav  and
      Villavicencio, Aline",
    booktitle = "Proceedings of the 60th Annual Meeting of the Association for Computational Linguistics (Volume 2: Short Papers)",
    month = may,
    year = "2022",
    address = "Dublin, Ireland",
    publisher = "Association for Computational Linguistics",
    url = "https://aclanthology.org/2022.acl-short.86/",
    doi = "10.18653/v1/2022.acl-short.86",
    pages = "762--770"
}

@inproceedings{reimers-gurevych-2019-sentence,
    title = "Sentence-{BERT}: Sentence Embeddings using {S}iamese {BERT}-Networks",
    author = "Reimers, Nils  and
      Gurevych, Iryna",
    editor = "Inui, Kentaro  and
      Jiang, Jing  and
      Ng, Vincent  and
      Wan, Xiaojun",
    booktitle = "Proceedings of the 2019 Conference on Empirical Methods in Natural Language Processing and the 9th International Joint Conference on Natural Language Processing (EMNLP-IJCNLP)",
    month = nov,
    year = "2019",
    address = "Hong Kong, China",
    publisher = "Association for Computational Linguistics",
    url = "https://aclanthology.org/D19-1410/",
    doi = "10.18653/v1/D19-1410",
    pages = "3982--3992"
}

@inproceedings{
wang2023selfconsistency,
title={Self-Consistency Improves Chain of Thought Reasoning in Language Models},
author={Xuezhi Wang and Jason Wei and Dale Schuurmans and Quoc V Le and Ed H. Chi and Sharan Narang and Aakanksha Chowdhery and Denny Zhou},
booktitle={The Eleventh International Conference on Learning Representations },
year={2023},
url={https://openreview.net/forum?id=1PL1NIMMrw}
}

@misc{cobbe2021trainingverifierssolvemath,
      title={Training Verifiers to Solve Math Word Problems}, 
      author={Karl Cobbe and Vineet Kosaraju and Mohammad Bavarian and Mark Chen and Heewoo Jun and Lukasz Kaiser and Matthias Plappert and Jerry Tworek and Jacob Hilton and Reiichiro Nakano and Christopher Hesse and John Schulman},
      year={2021},
      eprint={2110.14168},
      archivePrefix={arXiv},
      primaryClass={cs.LG},
      url={https://arxiv.org/abs/2110.14168}, 
}

@misc{chen2021evaluatinglargelanguagemodels,
      title={Evaluating Large Language Models Trained on Code}, 
      author={Mark Chen and Jerry Tworek and Heewoo Jun and Qiming Yuan and Henrique Ponde de Oliveira Pinto and Jared Kaplan and Harri Edwards and Yuri Burda and Nicholas Joseph and Greg Brockman and Alex Ray and Raul Puri and Gretchen Krueger and Michael Petrov and Heidy Khlaaf and Girish Sastry and Pamela Mishkin and Brooke Chan and Scott Gray and Nick Ryder and Mikhail Pavlov and Alethea Power and Lukasz Kaiser and Mohammad Bavarian and Clemens Winter and Philippe Tillet and Felipe Petroski Such and Dave Cummings and Matthias Plappert and Fotios Chantzis and Elizabeth Barnes and Ariel Herbert-Voss and William Hebgen Guss and Alex Nichol and Alex Paino and Nikolas Tezak and Jie Tang and Igor Babuschkin and Suchir Balaji and Shantanu Jain and William Saunders and Christopher Hesse and Andrew N. Carr and Jan Leike and Josh Achiam and Vedant Misra and Evan Morikawa and Alec Radford and Matthew Knight and Miles Brundage and Mira Murati and Katie Mayer and Peter Welinder and Bob McGrew and Dario Amodei and Sam McCandlish and Ilya Sutskever and Wojciech Zaremba},
      year={2021},
      eprint={2107.03374},
      archivePrefix={arXiv},
      primaryClass={cs.LG},
      url={https://arxiv.org/abs/2107.03374}, 
}

@article{shao2024deepseekmath,
  title={Deepseekmath: Pushing the limits of mathematical reasoning in open language models},
  author={Shao, Zhihong and Wang, Peiyi and Zhu, Qihao and Xu, Runxin and Song, Junxiao and Bi, Xiao and Zhang, Haowei and Zhang, Mingchuan and Li, YK and Wu, Yang and others},
  journal={arXiv preprint arXiv:2402.03300},
  year={2024}
}

@inproceedings{
lambert2025tulu,
title={Tulu 3: Pushing Frontiers in Open Language Model Post-Training},
author={Nathan Lambert and Jacob Morrison and Valentina Pyatkin and Shengyi Huang and Hamish Ivison and Faeze Brahman and Lester James Validad Miranda and Alisa Liu and Nouha Dziri and Xinxi Lyu and Yuling Gu and Saumya Malik and Victoria Graf and Jena D. Hwang and Jiangjiang Yang and Ronan Le Bras and Oyvind Tafjord and Christopher Wilhelm and Luca Soldaini and Noah A. Smith and Yizhong Wang and Pradeep Dasigi and Hannaneh Hajishirzi},
booktitle={Second Conference on Language Modeling},
year={2025},
url={https://openreview.net/forum?id=i1uGbfHHpH}
}

@article{shumailov2024model,
  title={AI models collapse when trained on recursively generated data},
  author={Shumailov, Ilia and Shumaylov, Zakhar and Zhao, Yiren and Papernot, Nicolas and Anderson, Ross and Gal, Yarin},
  journal={Nature},
  volume={631},
  number={8022},
  pages={755--759},
  year={2024},
  publisher={Nature Publishing Group UK London}
}

@inproceedings{
dang2025diversity,
title={Assessing Diversity Collapse in Reasoning},
author={Xingyu Dang and Christina Baek and J Zico Kolter and Aditi Raghunathan},
booktitle={Scaling Self-Improving Foundation Models without Human Supervision},
year={2025},
url={https://openreview.net/forum?id=AMiKsHLjQh}
}

@inproceedings{
yue2025does,
title={Does Reinforcement Learning Really Incentivize Reasoning Capacity in {LLM}s Beyond the Base Model?},
author={Yang Yue and Zhiqi Chen and Rui Lu and Andrew Zhao and Zhaokai Wang and Yang Yue and Shiji Song and Gao Huang},
booktitle={The Thirty-ninth Annual Conference on Neural Information Processing Systems},
year={2025},
url={https://openreview.net/forum?id=4OsgYD7em5}
}

@inproceedings{
padmakumar2024writing,
title={Does Writing with Language Models Reduce Content Diversity?},
author={Vishakh Padmakumar and He He},
booktitle={The Twelfth International Conference on Learning Representations},
year={2024},
url={https://openreview.net/forum?id=Feiz5HtCD0}
}

@inproceedings{tevet-berant-2021-evaluating,
    title = "Evaluating the Evaluation of Diversity in Natural Language Generation",
    author = "Tevet, Guy  and
      Berant, Jonathan",
    editor = "Merlo, Paola  and
      Tiedemann, Jorg  and
      Tsarfaty, Reut",
    booktitle = "Proceedings of the 16th Conference of the European Chapter of the Association for Computational Linguistics: Main Volume",
    month = apr,
    year = "2021",
    address = "Online",
    publisher = "Association for Computational Linguistics",
    url = "https://aclanthology.org/2021.eacl-main.25/",
    doi = "10.18653/v1/2021.eacl-main.25",
    pages = "326--346"
}

@inproceedings{
verine2025improving,
title={Improving Diversity in Language Models: When Temperature Fails, Change the Loss},
author={Alexandre Verine and Florian Le Bronnec and Kunhao Zheng and Alexandre Allauzen and Yann Chevaleyre and benjamin negrevergne},
booktitle={Forty-second International Conference on Machine Learning},
year={2025},
url={https://openreview.net/forum?id=RsyMfsqzeG}
}

@inproceedings{
shypula2025evaluating,
title={Evaluating the Diversity and Quality of {LLM} Generated Content},
author={Alexander Shypula and Shuo Li and Botong Zhang and Vishakh Padmakumar and Kayo Yin and Osbert Bastani},
booktitle={Second Conference on Language Modeling},
year={2025},
url={https://openreview.net/forum?id=O7bF6nlSOD}
}

@article{guo-etal-2025-benchmarking-linguistic,
    title = "Benchmarking Linguistic Diversity of Large Language Models",
    author = "Guo, Yanzhu  and
      Shang, Guokan  and
      Clavel, Chlo{\'e}",
    journal = "Transactions of the Association for Computational Linguistics",
    volume = "13",
    year = "2025",
    address = "Cambridge, MA",
    publisher = "MIT Press",
    url = "https://aclanthology.org/2025.tacl-1.69/",
    doi = "10.1162/tacl.a.47",
    pages = "1507--1526"
}

@inproceedings{
li2025preserving,
title={Preserving Diversity in Supervised Fine-Tuning of Large Language Models},
author={Ziniu Li and Congliang Chen and Tian Xu and Zeyu Qin and Jiancong Xiao and Zhi-Quan Luo and Ruoyu Sun},
booktitle={The Thirteenth International Conference on Learning Representations},
year={2025},
url={https://openreview.net/forum?id=NQEe7B7bSw}
}

@misc{li2025darling,
      title={Jointly Reinforcing Diversity and Quality in Language Model Generations}, 
      author={Tianjian Li and Yiming Zhang and Ping Yu and Swarnadeep Saha and Daniel Khashabi and Jason Weston and Jack Lanchantin and Tianlu Wang},
      year={2025},
      eprint={2509.02534},
      archivePrefix={arXiv},
      primaryClass={cs.CL},
      url={https://arxiv.org/abs/2509.02534}, 
}

@inproceedings{
chakraborty2024maxmin,
title={MaxMin-{RLHF}: Alignment with Diverse Human Preferences},
author={Souradip Chakraborty and Jiahao Qiu and Hui Yuan and Alec Koppel and Dinesh Manocha and Furong Huang and Amrit Bedi and Mengdi Wang},
booktitle={Forty-first International Conference on Machine Learning},
year={2024},
url={https://openreview.net/forum?id=8tzjEMF0Vq}
}

@misc{peeperkorn2025mindgapconformativedecoding,
      title={Mind the Gap: Conformative Decoding to Improve Output Diversity of Instruction-Tuned Large Language Models}, 
      author={Max Peeperkorn and Tom Kouwenhoven and Dan Brown and Anna Jordanous},
      year={2025},
      eprint={2507.20956},
      archivePrefix={arXiv},
      primaryClass={cs.CL},
      url={https://arxiv.org/abs/2507.20956}, 
}

@inproceedings{
omahony2024attributing,
title={Attributing Mode Collapse in the fine-tuning of Large Language Models},
author={Laura O'Mahony and Leo Grinsztajn and Hailey Schoelkopf and Stella Biderman},
booktitle={ICLR 2024 Workshop on Mathematical and Empirical Understanding of Foundation Models},
year={2024},
url={https://openreview.net/forum?id=3pDMYjpOxk}
}

@inproceedings{
snell2025scaling,
title={Scaling {LLM} Test-Time Compute Optimally Can be More Effective than Scaling Parameters for Reasoning},
author={Charlie Victor Snell and Jaehoon Lee and Kelvin Xu and Aviral Kumar},
booktitle={The Thirteenth International Conference on Learning Representations},
year={2025},
url={https://openreview.net/forum?id=4FWAwZtd2n}
}

@article{xiao2024preference,
  title={On the algorithmic bias of aligning large language models with rlhf: Preference collapse and matching regularization},
  author={Xiao, Jiancong and Li, Ziniu and Xie, Xingyu and Getzen, Emily and Fang, Cong and Long, Qi and Su, Weijie J},
  journal={arXiv preprint arXiv:2405.16455},
  year={2024}
}

@inproceedings{fan-etal-2018-hierarchical,
    title = "Hierarchical Neural Story Generation",
    author = "Fan, Angela  and
      Lewis, Mike  and
      Dauphin, Yann",
    editor = "Gurevych, Iryna  and
      Miyao, Yusuke",
    booktitle = "Proceedings of the 56th Annual Meeting of the Association for Computational Linguistics (Volume 1: Long Papers)",
    month = jul,
    year = "2018",
    address = "Melbourne, Australia",
    publisher = "Association for Computational Linguistics",
    url = "https://aclanthology.org/P18-1082/",
    doi = "10.18653/v1/P18-1082",
    pages = "889--898"
}

@inproceedings{lin-etal-2022-truthfulqa,
    title = "{T}ruthful{QA}: Measuring How Models Mimic Human Falsehoods",
    author = "Lin, Stephanie  and
      Hilton, Jacob  and
      Evans, Owain",
    editor = "Muresan, Smaranda  and
      Nakov, Preslav  and
      Villavicencio, Aline",
    booktitle = "Proceedings of the 60th Annual Meeting of the Association for Computational Linguistics (Volume 1: Long Papers)",
    month = may,
    year = "2022",
    address = "Dublin, Ireland",
    publisher = "Association for Computational Linguistics",
    url = "https://aclanthology.org/2022.acl-long.229/",
    doi = "10.18653/v1/2022.acl-long.229",
    pages = "3214--3252"
}

@inproceedings{
hendrycks2021measuring,
title={Measuring Mathematical Problem Solving With the {MATH} Dataset},
author={Dan Hendrycks and Collin Burns and Saurav Kadavath and Akul Arora and Steven Basart and Eric Tang and Dawn Song and Jacob Steinhardt},
booktitle={Thirty-fifth Conference on Neural Information Processing Systems Datasets and Benchmarks Track (Round 2)},
year={2021},
url={https://openreview.net/forum?id=7Bywt2mQsCe}
}

@inproceedings{
lin2025wildbench,
title={WildBench: Benchmarking {LLM}s with Challenging Tasks from Real Users in the Wild},
author={Bill Yuchen Lin and Yuntian Deng and Khyathi Chandu and Abhilasha Ravichander and Valentina Pyatkin and Nouha Dziri and Ronan Le Bras and Yejin Choi},
booktitle={The Thirteenth International Conference on Learning Representations},
year={2025},
url={https://openreview.net/forum?id=MKEHCx25xp}
}

@inproceedings{
kirk2024prism,
title={The {PRISM} Alignment Dataset: What Participatory, Representative and Individualised Human Feedback Reveals About the Subjective and Multicultural Alignment of Large Language Models},
author={Hannah Rose Kirk and Alexander Whitefield and Paul R{\"o}ttger and Andrew Michael Bean and Katerina Margatina and Rafael Mosquera and Juan Manuel Ciro and Max Bartolo and Adina Williams and He He and Bertie Vidgen and Scott A. Hale},
booktitle={The Thirty-eight Conference on Neural Information Processing Systems Datasets and Benchmarks Track},
year={2024},
url={https://openreview.net/forum?id=DFr5hteojx}
}

@misc{zhou2023ifeval,
      title={Instruction-Following Evaluation for Large Language Models}, 
      author={Jeffrey Zhou and Tianjian Lu and Swaroop Mishra and Siddhartha Brahma and Sujoy Basu and Yi Luan and Denny Zhou and Le Hou},
      year={2023},
      eprint={2311.07911},
      archivePrefix={arXiv},
      primaryClass={cs.CL},
      url={https://arxiv.org/abs/2311.07911}, 
}

@misc{alpaca,
  author = {Rohan Taori and Ishaan Gulrajani and Tianyi Zhang and Yann Dubois and Xuechen Li and Carlos Guestrin and Percy Liang and Tatsunori B. Hashimoto },
  title = {Stanford Alpaca: An Instruction-following LLaMA model},
  year = {2023},
  publisher = {GitHub},
  journal = {GitHub repository},
  howpublished = {\url{https://github.com/tatsu-lab/stanford_alpaca}},
}

@misc{austin2021programsynthesislargelanguage,
      title={Program Synthesis with Large Language Models}, 
      author={Jacob Austin and Augustus Odena and Maxwell Nye and Maarten Bosma and Henryk Michalewski and David Dohan and Ellen Jiang and Carrie Cai and Michael Terry and Quoc Le and Charles Sutton},
      year={2021},
      eprint={2108.07732},
      archivePrefix={arXiv},
      primaryClass={cs.PL},
      url={https://arxiv.org/abs/2108.07732}, 
}

@inproceedings{volske-etal-2017-tl,
    title = "{TL};{DR}: Mining {R}eddit to Learn Automatic Summarization",
    author = {V{\"o}lske, Michael  and
      Potthast, Martin  and
      Syed, Shahbaz  and
      Stein, Benno},
    editor = "Wang, Lu  and
      Cheung, Jackie Chi Kit  and
      Carenini, Giuseppe  and
      Liu, Fei",
    booktitle = "Proceedings of the Workshop on New Frontiers in Summarization",
    month = sep,
    year = "2017",
    address = "Copenhagen, Denmark",
    publisher = "Association for Computational Linguistics",
    url = "https://aclanthology.org/W17-4508/",
    doi = "10.18653/v1/W17-4508",
    pages = "59--63"
}

@inproceedings{narayan-etal-2018-dont,
    title = "Don{'}t Give Me the Details, Just the Summary! Topic-Aware Convolutional Neural Networks for Extreme Summarization",
    author = "Narayan, Shashi  and
      Cohen, Shay B.  and
      Lapata, Mirella",
    editor = "Riloff, Ellen  and
      Chiang, David  and
      Hockenmaier, Julia  and
      Tsujii, Jun{'}ichi",
    booktitle = "Proceedings of the 2018 Conference on Empirical Methods in Natural Language Processing",
    month = oct # "-" # nov,
    year = "2018",
    address = "Brussels, Belgium",
    publisher = "Association for Computational Linguistics",
    url = "https://aclanthology.org/D18-1206/",
    doi = "10.18653/v1/D18-1206",
    pages = "1797--1807"
}

@inproceedings{nallapati-etal-2016-abstractive,
    title = "Abstractive Text Summarization using Sequence-to-sequence {RNN}s and Beyond",
    author = "Nallapati, Ramesh  and
      Zhou, Bowen  and
      dos Santos, Cicero  and
      Gu{\ensuremath{\dot{}}}l{\c{c}}ehre, {\c{C}}a{\u{g}}lar  and
      Xiang, Bing",
    editor = "Riezler, Stefan  and
      Goldberg, Yoav",
    booktitle = "Proceedings of the 20th {SIGNLL} Conference on Computational Natural Language Learning",
    month = aug,
    year = "2016",
    address = "Berlin, Germany",
    publisher = "Association for Computational Linguistics",
    url = "https://aclanthology.org/K16-1028/",
    doi = "10.18653/v1/K16-1028",
    pages = "280--290"
}

@article{
friedman2023the,
title={The Vendi Score: A Diversity Evaluation Metric for Machine Learning},
author={Dan Friedman and Adji Bousso Dieng},
journal={Transactions on Machine Learning Research},
issn={2835-8856},
year={2023},
url={https://openreview.net/forum?id=g97OHbQyk1},
note={}
}

@misc{jain2025task,
      title={LLM Output Homogenization is Task Dependent}, 
      author={Shomik Jain and Jack Lanchantin and Maximilian Nickel and Karen Ullrich and Ashia Wilson and Jamelle Watson-Daniels},
      year={2025},
      eprint={2509.21267},
      archivePrefix={arXiv},
      primaryClass={cs.CL},
      url={https://arxiv.org/abs/2509.21267}, 
}

@inproceedings{
jiang2025hivemind,
title={Artificial Hivemind: The Open-Ended Homogeneity of Language Models (and Beyond)},
author={Liwei Jiang and Yuanjun Chai and Margaret Li and Mickel Liu and Raymond Fok and Nouha Dziri and Yulia Tsvetkov and Maarten Sap and Yejin Choi},
booktitle={The Thirty-ninth Annual Conference on Neural Information Processing Systems Datasets and Benchmarks Track},
year={2025},
url={https://openreview.net/forum?id=saDOrrnNTz}
}

@inproceedings{kamigaito2025diversity,
    title = "Diversity Explains Inference Scaling Laws: Through a Case Study of Minimum {B}ayes Risk Decoding",
    author = "Kamigaito, Hidetaka  and
      Deguchi, Hiroyuki  and
      Sakai, Yusuke  and
      Hayashi, Katsuhiko  and
      Watanabe, Taro",
    editor = "Che, Wanxiang  and
      Nabende, Joyce  and
      Shutova, Ekaterina  and
      Pilehvar, Mohammad Taher",
    booktitle = "Proceedings of the 63rd Annual Meeting of the Association for Computational Linguistics (Volume 1: Long Papers)",
    month = jul,
    year = "2025",
    address = "Vienna, Austria",
    publisher = "Association for Computational Linguistics",
    url = "https://aclanthology.org/2025.acl-long.1410/",
    doi = "10.18653/v1/2025.acl-long.1410",
    pages = "29060--29094",
    ISBN = "979-8-89176-251-0"
}

@misc{wright2025epistemic,
      title={Epistemic Diversity and Knowledge Collapse in Large Language Models}, 
      author={Dustin Wright and Sarah Masud and Jared Moore and Srishti Yadav and Maria Antoniak and Peter Ebert Christensen and Chan Young Park and Isabelle Augenstein},
      year={2026},
      eprint={2510.04226},
      archivePrefix={arXiv},
      primaryClass={cs.CL},
      url={https://arxiv.org/abs/2510.04226}, 
}

@inproceedings{
wei2022chain,
title={Chain of Thought Prompting Elicits Reasoning in Large Language Models},
author={Jason Wei and Xuezhi Wang and Dale Schuurmans and Maarten Bosma and brian ichter and Fei Xia and Ed H. Chi and Quoc V Le and Denny Zhou},
booktitle={Advances in Neural Information Processing Systems},
editor={Alice H. Oh and Alekh Agarwal and Danielle Belgrave and Kyunghyun Cho},
year={2022},
url={https://openreview.net/forum?id=_VjQlMeSB_J}
}

@inproceedings{anderson2024homogenization,
   title={Homogenization Effects of Large Language Models on Human Creative Ideation},
   url={http://dx.doi.org/10.1145/3635636.3656204},
   DOI={10.1145/3635636.3656204},
   booktitle={Creativity and Cognition},
   publisher={ACM},
   author={Anderson, Barrett R and Shah, Jash Hemant and Kreminski, Max},
   year={2024},
   month=jun, pages={413–425},
   collection={C&C ’24} }

@inproceedings{
west2025base,
title={Base Models Beat Aligned Models at Randomness and Creativity},
author={Peter West and Christopher Potts},
booktitle={Second Conference on Language Modeling},
year={2025},
url={https://openreview.net/forum?id=vqN8uom4A1}
}

@inproceedings{lewislim2025cot,
    title = "Analysing Chain of Thought Dynamics: Active Guidance or Unfaithful Post-hoc Rationalisation?",
    author = "Lewis-Lim, Samuel  and
      Tan, Xingwei  and
      Zhao, Zhixue  and
      Aletras, Nikolaos",
    editor = "Christodoulopoulos, Christos  and
      Chakraborty, Tanmoy  and
      Rose, Carolyn  and
      Peng, Violet",
    booktitle = "Proceedings of the 2025 Conference on Empirical Methods in Natural Language Processing",
    month = nov,
    year = "2025",
    address = "Suzhou, China",
    publisher = "Association for Computational Linguistics",
    url = "https://aclanthology.org/2025.emnlp-main.1516/",
    doi = "10.18653/v1/2025.emnlp-main.1516",
    pages = "29838--29853",
    ISBN = "979-8-89176-332-6"
}

@inproceedings{
geng2025deltalearning,
title={The Delta Learning Hypothesis: Preference Tuning on Weak Data can Yield Strong Gains},
author={Scott Geng and Hamish Ivison and Chun-Liang Li and Maarten Sap and Jerry Li and Ranjay Krishna and Pang Wei Koh},
booktitle={Second Conference on Language Modeling},
year={2025},
url={https://openreview.net/forum?id=9rwtezthwo}
}

@misc{qwq32b,
  title={Qwq: Reflect deeply on the boundaries of the unknown},
  author={Team, Qwen},
  year={2024}
}

@misc{ma2025reasoning,
      title={Reasoning Models Can Be Effective Without Thinking}, 
      author={Wenjie Ma and Jingxuan He and Charlie Snell and Tyler Griggs and Sewon Min and Matei Zaharia},
      year={2025},
      eprint={2504.09858},
      archivePrefix={arXiv},
      primaryClass={cs.AI},
      url={https://arxiv.org/abs/2504.09858}, 
}

@misc{lighteval,
  author = {Habib, Nathan and Fourrier, Clémentine and Kydlíček, Hynek and Wolf, Thomas and Tunstall, Lewis},
  title = {LightEval: A lightweight framework for LLM evaluation},
  year = {2023},
  version = {0.11.0},
  url = {https://github.com/huggingface/lighteval}
}

@misc{kwon2023efficient,
      title={Efficient Memory Management for Large Language Model Serving with PagedAttention}, 
      author={Woosuk Kwon and Zhuohan Li and Siyuan Zhuang and Ying Sheng and Lianmin Zheng and Cody Hao Yu and Joseph E. Gonzalez and Hao Zhang and Ion Stoica},
      year={2023},
      eprint={2309.06180},
      archivePrefix={arXiv},
      primaryClass={cs.LG},
      url={https://arxiv.org/abs/2309.06180}, 
}

@inproceedings{lake2025overton,
    title = "From Distributional to Overton Pluralism: Investigating Large Language Model Alignment",
    author = "Lake, Thom  and
      Choi, Eunsol  and
      Durrett, Greg",
    editor = "Chiruzzo, Luis  and
      Ritter, Alan  and
      Wang, Lu",
    booktitle = "Proceedings of the 2025 Conference of the Nations of the Americas Chapter of the Association for Computational Linguistics: Human Language Technologies (Volume 1: Long Papers)",
    month = apr,
    year = "2025",
    address = "Albuquerque, New Mexico",
    publisher = "Association for Computational Linguistics",
    url = "https://aclanthology.org/2025.naacl-long.346/",
    doi = "10.18653/v1/2025.naacl-long.346",
    pages = "6794--6814",
    ISBN = "979-8-89176-189-6"
}

@InProceedings{gu2024cruxeval,
  title = 	 {{CRUXE}val: A Benchmark for Code Reasoning, Understanding and Execution},
  author =       {Gu, Alex and Roziere, Baptiste and Leather, Hugh James and Solar-Lezama, Armando and Synnaeve, Gabriel and Wang, Sida},
  booktitle = 	 {Proceedings of the 41st International Conference on Machine Learning},
  pages = 	 {16568--16621},
  year = 	 {2024},
  editor = 	 {Salakhutdinov, Ruslan and Kolter, Zico and Heller, Katherine and Weller, Adrian and Oliver, Nuria and Scarlett, Jonathan and Berkenkamp, Felix},
  volume = 	 {235},
  series = 	 {Proceedings of Machine Learning Research},
  month = 	 {21--27 Jul},
  publisher =    {PMLR},
  url = 	 {https://proceedings.mlr.press/v235/gu24c.html}
}

@misc{openai2023gpt4,
      title={GPT-4 Technical Report}, 
      author={OpenAI and Josh Achiam and Steven Adler and Sandhini Agarwal and Lama Ahmad and Ilge Akkaya and Florencia Leoni Aleman and Diogo Almeida and Janko Altenschmidt and Sam Altman and Shyamal Anadkat and Red Avila and Igor Babuschkin and Suchir Balaji and Valerie Balcom and Paul Baltescu and Haiming Bao and Mohammad Bavarian and Jeff Belgum and Irwan Bello and Jake Berdine and Gabriel Bernadett-Shapiro and Christopher Berner and Lenny Bogdonoff and Oleg Boiko and Madelaine Boyd and Anna-Luisa Brakman and Greg Brockman and Tim Brooks and Miles Brundage and Kevin Button and Trevor Cai and Rosie Campbell and Andrew Cann and Brittany Carey and Chelsea Carlson and Rory Carmichael and Brooke Chan and Che Chang and Fotis Chantzis and Derek Chen and Sully Chen and Ruby Chen and Jason Chen and Mark Chen and Ben Chess and Chester Cho and Casey Chu and Hyung Won Chung and Dave Cummings and Jeremiah Currier and Yunxing Dai and Cory Decareaux and Thomas Degry and Noah Deutsch and Damien Deville and Arka Dhar and David Dohan and Steve Dowling and Sheila Dunning and Adrien Ecoffet and Atty Eleti and Tyna Eloundou and David Farhi and Liam Fedus and Niko Felix and Simón Posada Fishman and Juston Forte and Isabella Fulford and Leo Gao and Elie Georges and Christian Gibson and Vik Goel and Tarun Gogineni and Gabriel Goh and Rapha Gontijo-Lopes and Jonathan Gordon and Morgan Grafstein and Scott Gray and Ryan Greene and Joshua Gross and Shixiang Shane Gu and Yufei Guo and Chris Hallacy and Jesse Han and Jeff Harris and Yuchen He and Mike Heaton and Johannes Heidecke and Chris Hesse and Alan Hickey and Wade Hickey and Peter Hoeschele and Brandon Houghton and Kenny Hsu and Shengli Hu and Xin Hu and Joost Huizinga and Shantanu Jain and Shawn Jain and Joanne Jang and Angela Jiang and Roger Jiang and Haozhun Jin and Denny Jin and Shino Jomoto and Billie Jonn and Heewoo Jun and Tomer Kaftan and Łukasz Kaiser and Ali Kamali and Ingmar Kanitscheider and Nitish Shirish Keskar and Tabarak Khan and Logan Kilpatrick and Jong Wook Kim and Christina Kim and Yongjik Kim and Jan Hendrik Kirchner and Jamie Kiros and Matt Knight and Daniel Kokotajlo and Łukasz Kondraciuk and Andrew Kondrich and Aris Konstantinidis and Kyle Kosic and Gretchen Krueger and Vishal Kuo and Michael Lampe and Ikai Lan and Teddy Lee and Jan Leike and Jade Leung and Daniel Levy and Chak Ming Li and Rachel Lim and Molly Lin and Stephanie Lin and Mateusz Litwin and Theresa Lopez and Ryan Lowe and Patricia Lue and Anna Makanju and Kim Malfacini and Sam Manning and Todor Markov and Yaniv Markovski and Bianca Martin and Katie Mayer and Andrew Mayne and Bob McGrew and Scott Mayer McKinney and Christine McLeavey and Paul McMillan and Jake McNeil and David Medina and Aalok Mehta and Jacob Menick and Luke Metz and Andrey Mishchenko and Pamela Mishkin and Vinnie Monaco and Evan Morikawa and Daniel Mossing and Tong Mu and Mira Murati and Oleg Murk and David Mély and Ashvin Nair and Reiichiro Nakano and Rajeev Nayak and Arvind Neelakantan and Richard Ngo and Hyeonwoo Noh and Long Ouyang and Cullen O'Keefe and Jakub Pachocki and Alex Paino and Joe Palermo and Ashley Pantuliano and Giambattista Parascandolo and Joel Parish and Emy Parparita and Alex Passos and Mikhail Pavlov and Andrew Peng and Adam Perelman and Filipe de Avila Belbute Peres and Michael Petrov and Henrique Ponde de Oliveira Pinto and Michael and Pokorny and Michelle Pokrass and Vitchyr H. Pong and Tolly Powell and Alethea Power and Boris Power and Elizabeth Proehl and Raul Puri and Alec Radford and Jack Rae and Aditya Ramesh and Cameron Raymond and Francis Real and Kendra Rimbach and Carl Ross and Bob Rotsted and Henri Roussez and Nick Ryder and Mario Saltarelli and Ted Sanders and Shibani Santurkar and Girish Sastry and Heather Schmidt and David Schnurr and John Schulman and Daniel Selsam and Kyla Sheppard and Toki Sherbakov and Jessica Shieh and Sarah Shoker and Pranav Shyam and Szymon Sidor and Eric Sigler and Maddie Simens and Jordan Sitkin and Katarina Slama and Ian Sohl and Benjamin Sokolowsky and Yang Song and Natalie Staudacher and Felipe Petroski Such and Natalie Summers and Ilya Sutskever and Jie Tang and Nikolas Tezak and Madeleine B. Thompson and Phil Tillet and Amin Tootoonchian and Elizabeth Tseng and Preston Tuggle and Nick Turley and Jerry Tworek and Juan Felipe Cerón Uribe and Andrea Vallone and Arun Vijayvergiya and Chelsea Voss and Carroll Wainwright and Justin Jay Wang and Alvin Wang and Ben Wang and Jonathan Ward and Jason Wei and CJ Weinmann and Akila Welihinda and Peter Welinder and Jiayi Weng and Lilian Weng and Matt Wiethoff and Dave Willner and Clemens Winter and Samuel Wolrich and Hannah Wong and Lauren Workman and Sherwin Wu and Jeff Wu and Michael Wu and Kai Xiao and Tao Xu and Sarah Yoo and Kevin Yu and Qiming Yuan and Wojciech Zaremba and Rowan Zellers and Chong Zhang and Marvin Zhang and Shengjia Zhao and Tianhao Zheng and Juntang Zhuang and William Zhuk and Barret Zoph},
      year={2024},
      eprint={2303.08774},
      archivePrefix={arXiv},
      primaryClass={cs.CL},
      url={https://arxiv.org/abs/2303.08774}, 
}

@misc{liu2019robertarobustlyoptimizedbert,
      title={RoBERTa: A Robustly Optimized BERT Pretraining Approach}, 
      author={Yinhan Liu and Myle Ott and Naman Goyal and Jingfei Du and Mandar Joshi and Danqi Chen and Omer Levy and Mike Lewis and Luke Zettlemoyer and Veselin Stoyanov},
      year={2019},
      eprint={1907.11692},
      archivePrefix={arXiv},
      primaryClass={cs.CL},
      url={https://arxiv.org/abs/1907.11692}, 
}

@misc{ouyang2022traininglanguagemodelsfollow,
      title={Training language models to follow instructions with human feedback}, 
      author={Long Ouyang and Jeff Wu and Xu Jiang and Diogo Almeida and Carroll L. Wainwright and Pamela Mishkin and Chong Zhang and Sandhini Agarwal and Katarina Slama and Alex Ray and John Schulman and Jacob Hilton and Fraser Kelton and Luke Miller and Maddie Simens and Amanda Askell and Peter Welinder and Paul Christiano and Jan Leike and Ryan Lowe},
      year={2022},
      eprint={2203.02155},
      archivePrefix={arXiv},
      primaryClass={cs.CL},
      url={https://arxiv.org/abs/2203.02155}, 
}

@inproceedings{guo2022unixcoder,
    title = "{U}ni{X}coder: Unified Cross-Modal Pre-training for Code Representation",
    author = "Guo, Daya  and
      Lu, Shuai  and
      Duan, Nan  and
      Wang, Yanlin  and
      Zhou, Ming  and
      Yin, Jian",
    editor = "Muresan, Smaranda  and
      Nakov, Preslav  and
      Villavicencio, Aline",
    booktitle = "Proceedings of the 60th Annual Meeting of the Association for Computational Linguistics (Volume 1: Long Papers)",
    month = may,
    year = "2022",
    address = "Dublin, Ireland",
    publisher = "Association for Computational Linguistics",
    url = "https://aclanthology.org/2022.acl-long.499/",
    doi = "10.18653/v1/2022.acl-long.499",
    pages = "7212--7225"
}

@inproceedings{stasaski-hearst-2022-semantic,
    title = "Semantic Diversity in Dialogue with Natural Language Inference",
    author = "Stasaski, Katherine  and
      Hearst, Marti",
    editor = "Carpuat, Marine  and
      de Marneffe, Marie-Catherine  and
      Meza Ruiz, Ivan Vladimir",
    booktitle = "Proceedings of the 2022 Conference of the North American Chapter of the Association for Computational Linguistics: Human Language Technologies",
    month = jul,
    year = "2022",
    address = "Seattle, United States",
    publisher = "Association for Computational Linguistics",
    url = "https://aclanthology.org/2022.naacl-main.6/",
    doi = "10.18653/v1/2022.naacl-main.6",
    pages = "85--98"
}

@inproceedings{
zheng2023judging,
title={Judging {LLM}-as-a-Judge with {MT}-Bench and Chatbot Arena},
author={Lianmin Zheng and Wei-Lin Chiang and Ying Sheng and Siyuan Zhuang and Zhanghao Wu and Yonghao Zhuang and Zi Lin and Zhuohan Li and Dacheng Li and Eric Xing and Hao Zhang and Joseph E. Gonzalez and Ion Stoica},
booktitle={Thirty-seventh Conference on Neural Information Processing Systems Datasets and Benchmarks Track},
year={2023},
url={https://openreview.net/forum?id=uccHPGDlao}
}

@inproceedings{
li2024arenahard,
title={From Crowdsourced Data to High-quality Benchmarks: Arena-Hard and Benchbuilder Pipeline},
author={Tianle Li and Wei-Lin Chiang and Evan Frick and Lisa Dunlap and Tianhao Wu and Banghua Zhu and Joseph E. Gonzalez and Ion Stoica},
booktitle={Forty-second International Conference on Machine Learning},
year={2025},
url={https://openreview.net/forum?id=KfTf9vFvSn}
}

@inproceedings{lu2026rethinking,
    title = "Rethinking Creativity Evaluation: A Critical Analysis of Existing Creativity Evaluations",
    author = "Lu, Li-Chun  and
      Liu, Miri  and
      Lu, Pin Chun  and
      Tian, Yufei  and
      Sun, Shao-Hua  and
      Peng, Nanyun",
    editor = "Demberg, Vera  and
      Inui, Kentaro  and
      Marquez, Llu{\'i}s",
    booktitle = "Proceedings of the 19th Conference of the {E}uropean Chapter of the {A}ssociation for {C}omputational {L}inguistics (Volume 1: Long Papers)",
    month = mar,
    year = "2026",
    address = "Rabat, Morocco",
    publisher = "Association for Computational Linguistics",
    url = "https://aclanthology.org/2026.eacl-long.297/",
    doi = "10.18653/v1/2026.eacl-long.297",
    pages = "6329--6352",
    ISBN = "979-8-89176-380-7"
}

@misc{guha2025openthoughtsdatarecipesreasoning,
  title={OpenThoughts: Data Recipes for Reasoning Models}, 
  author={Etash Guha and Ryan Marten and Sedrick Keh and Negin Raoof and Georgios Smyrnis and Hritik Bansal and Marianna Nezhurina and Jean Mercat and Trung Vu and Zayne Sprague and Ashima Suvarna and Benjamin Feuer and Liangyu Chen and Zaid Khan and Eric Frankel and Sachin Grover and Caroline Choi and Niklas Muennighoff and Shiye Su and Wanjia Zhao and John Yang and Shreyas Pimpalgaonkar and Kartik Sharma and Charlie Cheng-Jie Ji and Yichuan Deng and Sarah Pratt and Vivek Ramanujan and Jon Saad-Falcon and Jeffrey Li and Achal Dave and Alon Albalak and Kushal Arora and Blake Wulfe and Chinmay Hegde and Greg Durrett and Sewoong Oh and Mohit Bansal and Saadia Gabriel and Aditya Grover and Kai-Wei Chang and Vaishaal Shankar and Aaron Gokaslan and Mike A. Merrill and Tatsunori Hashimoto and Yejin Choi and Jenia Jitsev and Reinhard Heckel and Maheswaran Sathiamoorthy and Alexandros G. Dimakis and Ludwig Schmidt},
  year={2025},
  eprint={2506.04178},
  archivePrefix={arXiv},
  primaryClass={cs.LG},
  url={https://arxiv.org/abs/2506.04178}, 
}

@article{toshniwal2024openmath2,
  title   = {OpenMathInstruct-2: Accelerating AI for Math with Massive Open-Source Instruction Data},
  author  = {Shubham Toshniwal and Wei Du and Ivan Moshkov and  Branislav Kisacanin and Alexan Ayrapetyan and Igor Gitman},
  year    = {2024},
  journal = {arXiv preprint arXiv:2410.01560}
}

@inproceedings{
  zhao2024wildchat,
  title={WildChat: 1M Chat{GPT} Interaction Logs in the Wild},
  author={Wenting Zhao and Xiang Ren and Jack Hessel and Claire Cardie and Yejin Choi and Yuntian Deng},
  booktitle={The Twelfth International Conference on Learning Representations},
  year={2024},
  url={https://openreview.net/forum?id=Bl8u7ZRlbM}
}

@misc{nvidia_nemotron_nano_v3_2025,
  title  = {{Nemotron 3 Nano}: Open, Efficient Mixture-of-Experts Hybrid {Mamba}-{Transformer} Model for {Agentic} Reasoning},
  author = {{NVIDIA}},
  year   = {2025},
  url    = {https://research.nvidia.com/labs/nemotron/files/NVIDIA-Nemotron-3-Nano-Technical-Report.pdf},
  note   = {Technical report}
}

@inproceedings{
sprague2025to,
title={To CoT or not to CoT? Chain-of-thought helps mainly on math and symbolic reasoning},
author={Zayne Rea Sprague and Fangcong Yin and Juan Diego Rodriguez and Dongwei Jiang and Manya Wadhwa and Prasann Singhal and Xinyu Zhao and Xi Ye and Kyle Mahowald and Greg Durrett},
booktitle={The Thirteenth International Conference on Learning Representations},
year={2025},
url={https://openreview.net/forum?id=w6nlcS8Kkn}
}
